**RESEARCH ARTICLE**

# Privacy and Accuracy Implications of Model Complexity and Integration in Heterogeneous Federated Learning

GERGELY D. NÉMETH[1], MIGUEL ÁNGEL LOZANO[2], NOVI QUADRIANTO[3,4], AND NURIA OLIVER[1], (Fellow, IEEE)

[1]ELLIS Alicante, 03005 Alicante, Spain
[2]University of Alicante, 03690 Alicante, Spain
[3]Department of Computer Science and Artificial Intelligence, University of Sussex, BN1 9RH Brighton, U.K.
[4]Department of Informatics, Basque Center for Applied Mathematics, 48009 Bilbao, Spain

Corresponding author: Gergely D. Németh (gergely@ellisalicante.org)

The work of Gergely D. Németh was supported in part by ELLIS Unit Alicante Foundation by European Commission under the Horizon Europe Program under Grant 101120237—ELIAS; in part by the Nominal Grant from the Regional Government of Valencia in Spain (Convenio Singular Signed with Generalitat Valenciana, Conselleria de Innovación, Industria, Comercio y Turismo, Dirección General de Innovación); in part by the Banco Sabadell Foundation; in part by European Union's Horizon 2020 Research and Innovation Program under ELISE under Grant 951847; and in part by European Research Council (ERC) Starting Grant for the Project "Bayesian Models and Algorithms for Fairness and Transparency," funded under European Union's Horizon 2020 Framework Program under Grant 851538. The work of Novi Quadrianto was supported in part by ERC Starting Grant for the Project "Bayesian Models and Algorithms for Fairness and Transparency," funded under European Union's Horizon 2020 Framework Program under Grant 851538. The work of Nuria Oliver was supported in part by ELLIS Unit Alicante Foundation by European Commission under the Horizon Europe Program under Grant 101120237—ELIAS; in part by the Nominal Grant from the Regional Government of Valencia in Spain (Convenio Singular Signed with Generalitat Valenciana, Conselleria de Innovación, Industria, Comercio y Turismo, Dirección General de Innovación).

**ABSTRACT** Federated Learning (FL) has been proposed as a privacy-preserving solution for distributed machine learning, particularly in *heterogeneous FL* settings where clients have varying computational capabilities and thus train models with different complexities compared to the server's model. However, FL is not without vulnerabilities: recent studies have shown that it is susceptible to membership inference attacks (MIA), which can compromise the privacy of client data. In this paper, we examine the intersection of these two aspects, heterogeneous FL and its privacy vulnerabilities, by focusing on the role of client model integration, the process through which the server integrates parameters from clients' smaller models into its larger model. To better understand this process, we first propose a **taxonomy** that categorizes existing heterogeneous FL methods and enables the design of **seven novel heterogeneous FL model integration** strategies. Using CIFAR-10, CIFAR-100, and FEMNIST vision datasets, we evaluate the privacy and accuracy trade-offs of these approaches under three types of MIAs. Our findings reveal **significant differences** in privacy leakage and performance depending on the integration method. Notably, introducing **randomness** in the model integration process enhances client privacy while maintaining competitive accuracy for both the clients and the server. This work provides quantitative light on the privacy-accuracy implications client model integration in heterogeneous FL settings, paving the way towards more secure and efficient FL systems.

The associate editor coordinating the review of this manuscript and approving it for publication was Li Zhang.

## I. INTRODUCTION
Deep neural networks require access to large amounts of training data to achieve competitive performance. This data dependency raises concerns regarding the safeguarding of sensitive information that might be encapsulated in the data.

Federated Learning (FL) has been proposed as a potential solution to mitigate such concerns [1], [2]. FL consists of a distributed machine learning approach that enables training models without the need to transfer the raw data from different devices or locations (clients) to a central server. In each iteration of the learning process, the server shares the parameters of the learned global model with the clients which perform local computations on their respective data to update their local parameters. Their updated model parameters are then sent back to the server, which aggregates the changes made by the clients to improve the global model.

Heterogeneous Federated Learning [3] refers to a more complex and realistic variant of FL where the participating clients have diverse conditions in terms of data, computing resources and model architectures. In this paper we focus on one type of heterogeneity namely *model size heterogeneity*, where different clients learn models of the same type but with varying complexities to adapt to their data and computational capabilities.[1]

Several approaches have been proposed in the literature to implement FL with model size heterogeneity [4], [5], [6], [7]. However, they are generally seen as independent methods.

Furthermore, FL is regarded as a privacy-preserving solution by design, given that the raw data never leaves the clients and only the model parameters are shared with the central server, which is of great importance in many practical scenarios, including healthcare [8], [9] and finance [10], [11], or intelligent smartphone interfaces [12]. However, recent work has shown that sensitive information about the original data can be inferred by analyzing the model parameters that are shared in the communication rounds [13], [14].

To the best of our knowledge, no study has explored the privacy implications of different heterogeneous FL methods. In this paper, we fill this gap by providing the following 4 contributions:

(1) We are the first to frame existing heterogeneous FL methods in a novel taxonomy according to the adopted strategies to integrate the clients' models in the server's model;

(2) This taxonomy leads to the identification of seven new heterogeneous FL methods to perform client model integration in the server;

(3) We empirically evaluate the 7 proposed heterogeneous FL approaches and 2 state-of-the-art methods –namely HeteroFL and FDropout– from the perspective of server accuracy, and client accuracy and privacy on three widely used image datasets;

(4) We find that randomness in the strategy used to perform client model integration enhances the clients' privacy while keeping competitive performance on the server's side. In sum, our work provides the first empirical evidence on the privacy-accuracy implications of client model integration in heterogeneous FL.

## II. RELATED WORK
In this section, we present the most relevant previous work on FL with model heterogeneity and on privacy attacks in FL.

### A. MODEL HETEROGENEITY IN FL
In traditional FL all clients use the same model architecture as the server. However, this approach is unrealistic when clients have different computational and communication capabilities. FL with heterogeneous client models has been proposed to address this limitation as it enables training a diversity of models in the clients according to their capacities. There are two broad types of heterogeneous FL methods:

#### 1) KNOWLEDGE TRANSFER
In the first category, clients leverage a public dataset to communicate via knowledge distillation, and learn different models without sharing a global model with the server [15], [16], [17]. While this design enables clients to train different model architectures without limitations, its disadvantage is the lack of a competitive model in the server.

#### 2) MODEL SIZE HETEROGENEITY
The second category consists methods with partial architecture sharing. For example, resource-restricted clients can learn a less complex model which is a smaller version of the server's model. In this case, both the server and client-side models are trained as part of the federation [4], [5], [18]. In the context of deep neural networks, the model compression on the clients side can be achieved by training models with fewer [18] or with simpler [4], [5], [6], [7], [19] layers. Our work focuses on heterogeneous FL methods in this category.

### B. MEMBERSHIP INFERENCE ATTACKS IN FL
While FL was initially motivated by the desire to preserve client data, recent studies have revealed that federated systems remain vulnerable to privacy attacks, specifically in the form of membership attacks [20], [21], [22], [23], [24]. To tackle this limitation, several privacy-preserving approaches for FL have been proposed to date, including local differential privacy [20], [21] and data augmentation [22], [23]. In our work, we focus on membership inference attacks (MIAs) and their implications on heterogeneous FL. In MIAs, the attacker's goal is to determine whether an individual data point was part of the dataset used to train the target model. While MIAs expose less private information than other attacks, such as memorization attacks, they are still of great concern as they constitute a confidentiality violation [25]. Membership inference can also be used as a building block for mounting extraction attacks for existing machine learning as a service systems [14]. Several types of MIAs have been proposed in the literature [26], [27]. In this work, we focus on black-box attacks where the attacker does not have full access to the models but is able to query them, which is a more realistic scenario than white-box attacks that assume full access to the models. We analyze the impact of three popular MIAs that use complementary strategies and hence offer a comprehensive evaluation of client privacy vulnerabilities in heterogeneous FL settings. Namely, the Yeom [28], LiRA [29], and tMIA [30] attacks. The Yeom attack is a simple, yet effective loss-based attack; LiRA is a good representative of shadow model-based techniques; and tMIA is a state-of-the-art knowledge distillation-based method to approximate the inspected model.

---
[1]In the following, we will use the term *heterogenous FL* to refer to heterogenous FL methods where the clients learn models of the same architecture but different complexities than the server's model.

While it is known that the larger the complexity of a model, the higher its vulnerability against MIAs [9], [28], as illustrated by Figure 4 and its corresponding section in the Appendix, we are not aware of any study of the impact on privacy of the model integration strategy adopted by the server in a heterogeneous FL setting.

## III. A TAXONOMY OF HETEROGENEOUS FL METHODS

In this section, propose a novel taxonomy of heterogenous FL methods which allows to both compare existing methods and identify seven new methods.

**TABLE 1.** Two broad groups of FL methods with model size heterogeneity. a) FL methods with dynamic selection of the clients' model size. All clients are assumed to hold a model of the same complexity as the server's model. These methods are not applicable to settings where clients have data and computation constraints, as is our case. Thus, they are beyond the scope of this paper. b) FL methods where the clients have a fixed model size, which can be smaller than the server's model size. In this case, the clients apply different strategies to select the portions from the server's model to be used in their training. The blue font corresponds to the newly proposed methods in this paper.

(a)

| Update | Dynamic client size selection methods | |
|---|---|---|
| | Random | Gradient |
| Each round | | Flado [7] |
| Each batch | FjORD [6] | |

(b)

| Coverage | Selection strategy | |
|---|---|---|
| | Resampled (S) | Fixed (F) |
| One group (O) | OSM, OSR | HeteroFL [5], OFR |
| Several groups (G) | GSR | GFM, GFR |
| Unique (U) | FDropout [4] | UFR |

### A. FORMULATION

We assume a heterogenous FL architecture in a computer vision task, where both the server and clients' models are Convolutional Neural Networks (CNNs) with a different number of channels in each layer, but the same number of layers.

Let $\theta$ denote the model parameters of the server's CNN, composed of $L$ layers represented by a weight matrix $A^{N \times M, l} \in \theta$ at each layer $l$. In such a setting, model reduction $\theta_c \subset \theta$ in client $c$ is achieved by limiting the size of each layer $l$ in the client's CNN according to the following principle: a layer in the server represented by weight matrix $A^{N \times M, l} \in \theta$ is reduced to size $N_c \times M_c$, where $N_c < N$ and $M_c < M$ such that every cell $a_c^{(i_c, j_c), l}$ in the reduced matrix $A_c^{N_c \times M_c, l}$ corresponds to a cell $a^{(i,j), l}$ in the server's matrix $A^{N \times M, l}$:

$$\forall i_c, j_c, l : a_c^{(i_c, j_c), l} \in A_c^l,$$
$$\exists i, j : a^{(i,j), l} \in A, a_c^{(i_c, j_c), l} = a^{(i, j), l} \quad (1)$$

In this scenario, there are two broad sets of methods proposed in the literature to perform client model integration in the server, as reflected in Table 1. The first group of methods, shown in Table 1(a), corresponds to algorithms that *dynamically* select the size of the clients' models but where all the clients are able to hold models of the same size as the server's model. Thus, these methods define an $Ms(\cdot)$ function that determines the $N_c^l \times M_c^l$ dimensions of the client's $c$ model to be used of the weight matrix $A_c^l$ in layer $l$ of their model $\theta_c$. Popular approaches in this category include Flado [7] and FjORD [6]. Note that they are not applicable to settings where clients have data and computation constraints, as it is our case. Hence, they are out-of-the-scope of this paper.

The second group, depicted in Table 1(b) and illustrated in Figure 1, corresponds to methods where the clients have *fixed-size* models that are typically smaller than the server's model. As we are considering CNNs, we refer to this family of methods as *channel selection* methods. The weights of each layer $l$ in a 2D CNN are defined by an $(N, M, H, W)^l$ dimensional tensor, where $M$ and $N$ are the input and output channels of the layer and $H$ and $W$ are the height and width of the kernel, respectively. Thus, $A^{N \times M, l}$ denotes the weight matrix of each linear layer $l$ in the server's model, where $M$ and $N$ are the input and output data dimensions [31], and $a^{(i,j), l}$ represents the kernel weights of the (i,j) position. In this case, $Ch(\cdot) : A^{N \times M, l} \to A_c^{N_c \times M_c, l}$ determines the mapping between the cells of the server's weight matrix $A^l$ and the client's $c$ smaller matrix $A_c^l$ for each layer $l$. Without a loss of generalization, we assume that the channels are sorted.

### B. A TAXONOMY OF HETEROGENEOUS FL METHODS

Figure 1 illustrates the proposed taxonomy of heterogeneous FL methods, according to three dimensions that characterize how the clients' models are integrated into the server's model.

#### 1) CLIENT GROUPING

The first dimension of the taxonomy refers to **client grouping**, classifying the methods in three classes: *one group* (O); *several groups* (G); and *unique* (U), depending on the number of channel sets used to train the models in the clients with smaller models than the server's model. In *one group*, all the clients use the same set of channels. In *several groups*, clients are clustered in groups such that clients in the same group use the same set of channels (Figure 1a) shows an example with 4 groups). *Unique* corresponds to federations where every client has their set of channels selected individually, illustrated by the rectangles with different colored patterns in Figure 1(a).

### 2) DYNAMICS

The second dimension (Figure 1b) characterizes the **dynamics** of the channel selection approach and defines two types: *fixed* (F) methods when the channel sets are defined at the beginning of the training, and *resampled* (S) methods when the channel sets are selected in each training round, $t_1 \ldots t_N$. The Figure illustrates the dynamics with 4 clients and in three training rounds $t_1$ to $t_3$. The clients' models are represented by rectangles with different colored patterns which represent the selection of channels from the server's model. In the *fixed* case, the clients get a variation of the same channels from the server in every round of training (hence the patterns in the rectangles are the same in the different training rounds), while when the channel sets are *resampled*, they get a new set of channels from the server every round, and hence the patterns change in each training round $t_i$.

### 3) POLICY

Finally, the third dimension (Figure 2c) concerns the **policy** for channel integration of the clients' models in the server's model and consists of two kinds: *submatrix* (M) methods if the selected channels are groups of adjacent channels and *random* (R) methods if the channels are selected randomly. In the Figure, with the submatrix policy the models from each of the 4 clients are integrated in non-overlapping sections of the server's matrix whereas with the random policy different portions of the clients' models are integrated in different sections of the server's model.

### C. EXISTING HETEROGENOUS FL ALGORITHMS

The proposed taxonomy enables us to characterize existing methods in heterogeneous FL.

#### 1) FDropout

In `FDropout` [4], the clients learn a CNN with the same architecture but fewer parameters (smaller weight matrices) than the server, and the server randomly drops a fixed number of units from each client [32], mapping the sparse model to a dense, smaller network by removing the dropped weights.

While the original formulation of `FDropout` used the same model size in all the clients, an extended variation was proposed by [6] that allows clients to have different model sizes and hence falls within the heterogenous FL definition used in this paper. In this variation, in each layer $l$ of the server's model, randomly selected cells, $a^{i,j,l}$ and their associated rows $i$ and columns $j$ are dropped from the weight matrix. The size of the client's model weight matrix in each layer can be set by the number of dropped rows and columns: $|Drop(N_l, N_{c,l})| = N_l - N_{c,l}$ and $|Drop(M_l, M_{c,l})| = M_l - M_{c,l}$, where $Drop(n, k)$ selects $k$ elements from $n$ randomly. Therefore, for each layer $l$ of the server's model, in `FDropout`:

$$a^{i,j,l} \in A_c : i \notin Drop(N_l, N_{c,l}), j \notin Drop(M_l, M_{c,l}). \quad (2)$$

According to our taxonomy, `FDropout` corresponds to a `USR` method because each client has a unique (U), random (R) set of channels that are resampled (S) in each training round.

#### 2) HeteroFL

`HeteroFL` [5] follows a similar idea as `FDropout` but with two key differences when selecting the channels in the clients with smaller models than the server: 1) all the clients learn from the same portion of the server's model; and 2) instead of randomly dropping cells, the clients always keep the top-left subset of the server's weight matrix for each layer in the network. Thus, in `HeteroFL`, the weight matrix $A_c^l$ of size $N_c \times M_c$ in layer $l$ and client $c$ corresponds to the top-left sub-matrix of the server's weight matrix $A^l$ of size $N \times M$:

$$\forall a_c^{i,j} \in A_c, a^{i,j} \in A : a_c^{i,j} = a^{i,j}, i = 1..N_c, j = 1..M_c. \quad (3)$$

According to our taxonomy, `HeteroFL` corresponds to an `OFM` method as there is only one client group (O) with fixed channels (F) that correspond to a sub-matrix (M) of the server's weight matrix.

### D. NEWLY PROPOSED HETEROGENEOUS FL METHODS

In addition to `HeteroFL` and `FDropout`, our taxonomy enables us to propose seven new methods for heterogeneous FL. In the following and for simplicity, we drop the superscript $l$ to denote the layer in the network.

1) **GFM:** In the `GFM` method, instead of selecting the top-left sub-matrix of the server's model (as in `HeteroFL`), the clients are randomly placed in N groups. In the following, we present the example where N = 4. Thus, the clients are assigned to one of 4 groups, $O, P, Q, R$. The client's channels are selected based on their group's policy, such that each cell from the original matrix is assigned to one cell in one of the four group.
    The matrix assigned to group $O$ is the same as the `HeteroFL` sub-matrix: it always selects the top-left cells of the server's matrix. Clients in group $R$ are assigned the bottom-right cells. The sub-matrices assigned to clients in groups $O$ and $P$ alternate between the bottom-left and the top-right cells. This is due to a restriction on how the input-output channels need to be connected. The top-right sub-matrix corresponds to selecting the second half of the input channels and the first half of the output channels. Therefore, if in layer $l$ the client selected the top-right sub-matrix, in the next layer it has to select one of the left sub-matrices, as they are the ones with the first half of the input channels. Note that this approach can be generalized to 9, 16,... groups, depending on the number of clients and the desired model size reduction. The cell assignment in the sub-matrices of each of the four groups is summarized in Equation 4 below.
    The top portion of Figure Figure 2 illustrates the GFM method. As seen in the Figure, the clients are grouped into four groups (G) with fixed channel sets (F) and

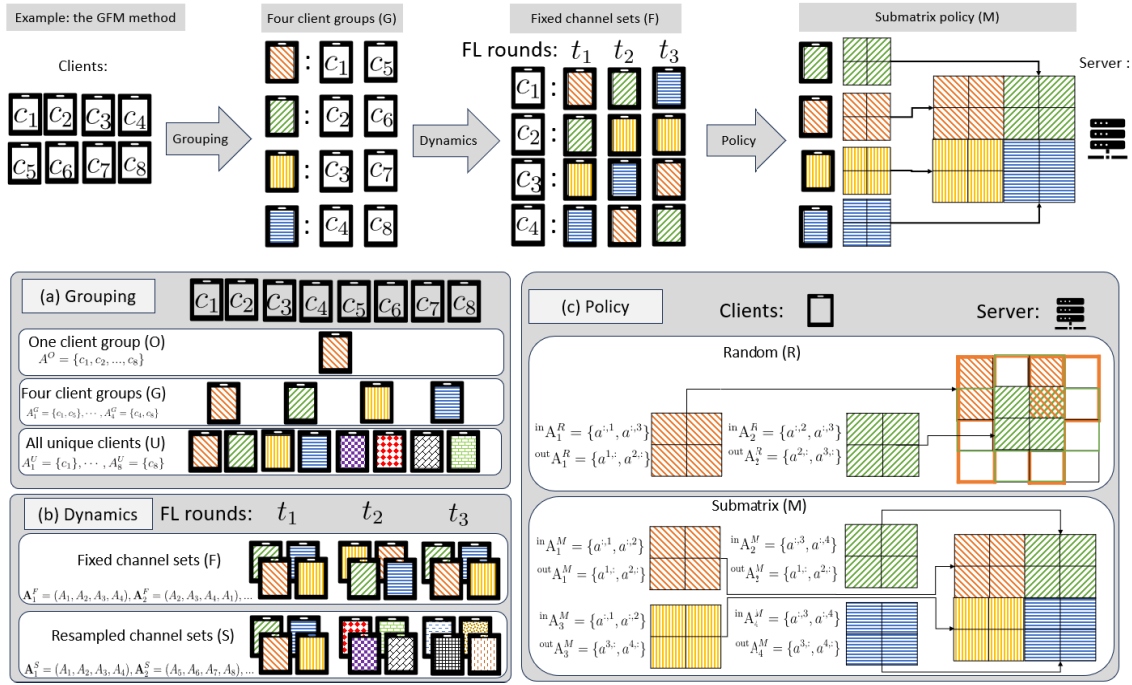

**FIGURE 1.** Proposed taxonomy of channel selection methods for heterogeneous FL architectures. The server and the clients learn the same type of models (e.g. CNNs) but with different numbers of units. The server selects which subset (channels in the case of a CNN) of its model is used to train the clients' models. We refer to this mechanism as *client model integration*. The taxonomy considers three dimensions: a) The groups of clients learning from the same server channels: one group (O), four groups (G), all unique clients (U); b) Dynamics in channel group selection: fixed at the beginning of the training (F), sampled in each round (S); and c) Policy for channel selection from the server's model: according to a submatrix structure (M), randomly (R). The top of the figure illustrates the taxonomy with one type of the proposed heterogeneous FL methods, namely GFM: there are four groups of clients (G) indicated by different colors, which use fixed channel sets (F) that are integrated in the server's model as submatrices (M).

integrating their models as submatrices of the server's model (M).

2) `GFR`: Compared to `GFM`, `GFR` differs in the set of channels in $A_O, A_P, A_Q,$ and $A_R$. Instead of selecting the first or the last $N_c$ and $M_c$ channels, the output channels are selected randomly, while the input channels match the output channels of the previous layer.

$$a^{(i,j),l} \in \begin{cases} A_O, & \text{if } 1 \leq i \leq N_c, 1 \leq j \leq M_c \\ A_P, & \text{if } 1 \leq i \leq N_c, M - M_c \leq j \leq M, l \text{ odd,} \\ & \text{or } N - N_c \leq i \leq N, 1 \leq j \leq M_c, l \text{ even} \\ A_Q, & \text{if } N - N_c \leq i \leq N, 1 \leq j \leq M_c, l \text{ odd,} \\ & \text{or } 1 \leq i \leq N_c, M - M_c \leq j \leq M, l \text{ even} \\ A_R, & \text{if } N - N_c \leq i \leq N, M - M_c \leq j \leq M \end{cases}$$
(4)

3) `GSR`: `GSR` is similar to `GFR` but in this case the set of channels are drawn randomly for each group in every round of training.

4) `OSM`: `OSM` generalizes `HeteroFL` by leveraging the channel sets $\{A_O, A_P, A_Q, A_R\}$ introduced in `GFM`, but in each training round all clients are using one of the 4 groups.

5) `OFR`: `OFR` is a variation of `HeteroFL` where instead of the top-left subset of channels in the server's weight matrix, the clients all get the same random set of channels for every round of training.

6) `OSR`: In `OSR`, the set of channels are drawn randomly in every training round, but all clients use the same set.

7) `UFR`: Finally, `UFR` selects $C$ unique sets of channels from the server's model which are defined at the beginning of the training and clients have access to one of the sets according to a new permutation every round. Therefore, in a federation with N clients, the clients receive the parameters from the same set of channels every N rounds.

We are interested in shedding light on the server accuracy and client accuracy-privacy trade-off of these 9 methods to perform client model integration in FL with heterogeneous models. Specifically, we focus on *membership inference attacks* or MIAs, as they represent a critical privacy threat in Federated Learning. MIAs allow adversaries to determine whether a particular data point was part of a client's training set by exploiting patterns in model updates or predictions. Given the diversity in model sizes in heterogeneous FL, the susceptibility of smaller, resource-constrained models to such attacks may differ from that of more complex models. By analyzing the performance of MIAs across methods to achieve heterogeneity in FL, we aim to understand the extent the model size in the client and the model integration strategy impact both privacy and accuracy. This analysis is crucial for developing robust FL frameworks that balance privacy guarantees and model performance in real-world settings with heterogeneous devices.

## IV. MEMBERSHIP INFERENCE ATTACKS IN FL

In Federated Learning, membership inference attacks can occur on the client or the server sides. In this paper, we focus on client attacks which occur when the attacker targets the client's model, $(f_c, \theta_c^t)$, for client $c = 1, \ldots, N$ in training round $t = 1, \ldots, T$. In a setting where all clients participate in the federation (stateful setting [33]), the attacker can collect a set of $k \leq T$ client updates $\Theta_c = \{\theta_c^{\tau_1}, \ldots, \theta_c^{\tau_k}\}$, $\tau_i \in \{1, .., T\}$; Specifically, we consider *client-side* attacks which take place on the last parameter update from the client to the server $\theta_c^T$ where the attacker aims to identify instances of the client's dataset $\mathbb{D}_c$ for client $c$.

Furthermore, we focus on *black-box* attacks, *i.e.* attacks where the attacker has no direct access to the model's parameters $\theta_g$ and architecture $f$, but it can query the model with data instances to get the model prediction $\hat{y}$. The attacker's purpose is to build an attacker model $\mathcal{A}$ that predicts, for data instance $(x, y)$, if it was part of the training data $\mathbb{D}_c$ of model $M(f, \theta_c, \mathbb{D}_c)$, for client $c = 1, \ldots, N$. Finally, we consider *passive* attacks where the attacker observes the behavior of a system without altering it.

Formally, the perfect attacker's model $\mathcal{A}$ is given by:

$$\mathcal{A}(f, \theta_c, (x, y)) = \begin{cases} 1, & \text{if } (x, y) \in \mathbb{D}_c, M(f, \theta_c, \mathbb{D}_c) \\ 0, & \text{otherwise.} \end{cases} \quad (5)$$

We study the performance of three different client-side, passive and black-box MIAs which are summarized below and are described in more detail in the Appendix.

- `Yeom` **attack**, where the attacker chooses a global threshold $\nu$, and selects every data instance with a loss lower than $\nu$ as a member of the training dataset in each client [28].
- `LiRA` **attack,** where the attacker has access to an auxiliary dataset $\mathbb{D}_a$ and trains shadow models $M_{sw}(f, \mathbb{D}_{sw})$ on random subsets of this dataset $\mathbb{D}_{sw} \subset \mathbb{D}_a$. The data instance is predicted to be a member of the client's training set if the target model's confidence score fits into the sample's confidence score distribution in the shadow models [29].
- `tMIA` **attack,** which leverages knowledge distillation to collect loss trajectories to identify member and non-member instances [30]. This attack builds on the idea that the snapshots of the loss after each training epoch (loss trajectory) can separate the member from non-member instances better than only using the final model's loss.

The selected attacks represent a distinct approaches to MIAs, ensuring a comprehensive evaluation and coverage of a variety of methodologies: (1) the `Yeom` attack is a simple, popular, loss-based and interpretable method; (2) `LiRA` is a good representative of shadow model-based techniques, which leverage synthetic data and advanced likelihood estimation methods to achieve high accuracy and scalability; and (3) `tMIA` is a state-of-the-art knowledge distillation-based method.

### A. PERFORMANCE METRICS OF MIAs
Three commonly used metrics to determine the performance of MIAs include:
- **AUC (Area under the curve):** This metric represents the area under the Receiver Operating Characteristic (ROC) curve, which plots the True Positive Rate (TPR) against the False Positive Rate (FPR) at various threshold settings. AUC ranges from 0.5 (random guessing) to 1.0 (perfect prediction). A higher AUC indicates better discrimination between members and non-members in the dataset and hence better performance of the MIA.
- **Attack advantage:** This measures how much better the attack model performs compared to random guessing. It is calculated as double the difference between the attack model's accuracy and 0.5 (the accuracy of a random classifier). A higher attack advantage indicates the attack is more effective at distinguishing members from non-members than a random model.
- **TPR@FPR 0.1:** This is the True Positive Rate (or recall) when the False Positive Rate is fixed at 0.1 (10%). It reflects the proportion of actual members that are correctly identified by the attack when only 10% of non-members are incorrectly classified as members. A higher TPR@FPR 0.1 indicates a more powerful attack with a low tolerance for false positives.

#### 1) PRIVACY-ACCURACY TRADE-OFF
Each of the heterogeneous FL methods is expected to provide a different privacy-accuracy trade-off, depending on how the client model integration is performed in the server. We formulate below three hypotheses that we empirically validate in our experiments.

#### a: FREQUENCY HYPOTHESIS (H1)
We hypothesize that heterogeneous FL methods where clients have access to the same set of channels more frequently perform better in terms of client-level accuracy but have a worse client-level privacy. For example, in `GFM` the clients access the same set of channels every four rounds. Thus, compared to the two state-of-the-art heterogeneous FL methods, namely `HeteroFL` (same set of channels per client across rounds) and `FDropout` (random selection of channels per client in each round), we expect this method to yield a client privacy-performance trade-off between these two existing methods.

Based on this hypothesis we expect:
- `OSR`, `GSR`, and `USR` (`FDropout`) to be the most resilient methods against MIAs but provide the worst client accuracy as the clients receive the parameters from a new set of channels in every round. Therefore, the same set is only repeated in every $\binom{N}{N_c}$ rounds on average for client $c$ with client channel size $N_c$ and server channel size $N$.
- `OFM` (`HeteroFL`) and `OFR` to be the most vulnerable against MIAs but achieve high client accuracies as the clients train using the parameters of the same set of channels in every round (1 round).

*b: SIMILARITY BETWEEN THE M AND R CATEGORIES (H2)*

In a CNN layer, as long as the selected input channels of layer $l$ match the output channels of layer $l-1$, the differences between variations $M$ and $R$ should be small. They differ only in the number of channels shared by client groups. We designed the sub-matrix category (M) to minimize the channel overlap between groups. Thus, we expect the models in the M and R categories to behave similarly regarding performance and privacy.

*c: THE DIFFERENCES IN THE PRIVACY-ACCURACY TRADE-OFF BETWEEN THE METHODS DECREASE AS THE NUMBER OF LARGE CLIENTS IN THE FEDERATION INCREASES (H3)*

The heterogeneous FL methods discussed in this paper are relevant when the majority of the clients learn smaller models than the server's model. Note that in cases when all the clients but one learn models of the same complexity as the server's model, the `UFR` and `OFR` methods become the same. Therefore, we expect the impact of the channel selection strategies to be larger when the majority of clients in the federation learn smaller models than the server's model.

We perform a comparative analysis of the proposed methods and empirically validate our hypotheses in experiments on commonly used vision datasets, as described next.

## V. EXPERIMENTS
### A. DATASETS
We perform experiments on two widely used image datasets: CIFAR-10 and CIFAR-100 given that, our work lies at the intersection of MIA techniques and heterogenous FL and, to the best of our knowledge, these two datasets are the only common datasets in the literature from both communities(FL: [1], [5], [6] and MIA: [22], [28]).

**CIFAR-10** [34] contains 60,000 images from 10 classes (50,000 images for training and validation and 10,000 images for testing).

**CIFAR-100** [34] has the same number of training and testing images as CIFAR-10 but with 100 classes and 500 training images per class.

We use a class-wise balanced, but client-wise weighted-distribution. We generate a data distribution using the Dirichlet distribution $Dir(\alpha)$ once, and apply the same split for each class. This ensures that each client has the same number of images from each class while they have different dataset sizes. The dataset size imbalance is controlled by the $\alpha \in (0, \infty)$ value: the larger the $\alpha$, the closer the allocation of training data to the uniform distribution and hence the closer to an IID scenario. Using $\alpha = 0.85$ this distribution generates clients with dataset size typically ranging from 1,000 to 10,000 samples. We apply *random crop* and *random flip* augmentations.

### B. METHODOLOGY
#### 1) MACHINE LEARNING MODEL
Given the nature of the data (images), we use a sequential CNN architecture, with convolutional, batch normalization and fully connected layers with trainable weights following [5]. We control the model complexity by changing the number of channels in the convolutional layers and the number of units in the final fully connected layer. In our experiments, we increase the complexity by factors of 2: each increase in the level of model complexity entails doubling the input and output channel sizes in each inner convolutional layer and the number of units in the final fully connected layer. A detailed description of the model architecture is available in the Appendix.

**TABLE 2.** Correlations of the three performance attack metrics (AUC, Adv, TPR@FPR 0.1%) on the three MIAs (`Yeom`, `LiRA`, `tMIA`) on the CIFAR-10 dataset. Note how AUC has the largest correlation across the three MIAs. Hence, we use in the experiments the average AUC over the three MIAs as the privacy performance metric.

| | Privacy Attack Metrics: AUC / Adv / TPR@FPR 0.1% | | |
|---|---|---|---|
| | `tMIA` | `LiRA` | `Yeom` |
| `tMIA` | **1.00** / 1.00 / 1.00 | **0.83** / 0.70 / 0.00 | **0.98** / 0.78 / 0.46 |
| `LiRA` | **0.83** / 0.70 / 0.00 | **1.00** / 1.00 / 1.00 | **0.85** / 0.68 / 0.02 |
| `Yeom` | **0.98** / 0.78 / 0.46 | **0.85** / 0.68 / 0.02 | **1.00** / 1.00 / 1.00 |

#### 2) EXPERIMENTAL SETUP
In all experiments we define a heterogeneous FL architecture with 10 clients which are trained with the Adam optimizer, a learning rate of 0.001 for one local epoch, a batch size of 128, and 150 rounds of FL. Experiments are repeated 3 times and we report mean values and standard deviation. The server learns a *large* model, which corresponds to a CNN network with 100k parameters. The clients learn a model with either the same complexity as the server's model or one complexity level below with 30k parameters (*small* model). All models are built in PyTorch [35] with the Flower federated framework [36]. FL clients are simulated in parallel on 2 AMD EPYC 7643 48-Core CPUs with 252GB RAM.

We train the 9 heterogeneous FL methods in a FL architecture with 10 clients, of which 2, 5, or 8 clients learn small models and the rest learn models of the same complexity as the server's model. Clients with a smaller dataset size are selected first to learn smaller models.

We also train as baselines two `FedAvg` baselines, FedAvg30k and FedAvg100k, where the server and the clients learn models with 30k and 100k parameters respectively. Thus, we train 29 different FL models for each data distribution.

While we do not perform experiments with heterogeneous FL architectures that include more than two levels of model complexity, we expect our results to extrapolate to other configurations in a similar way as reported in [5].

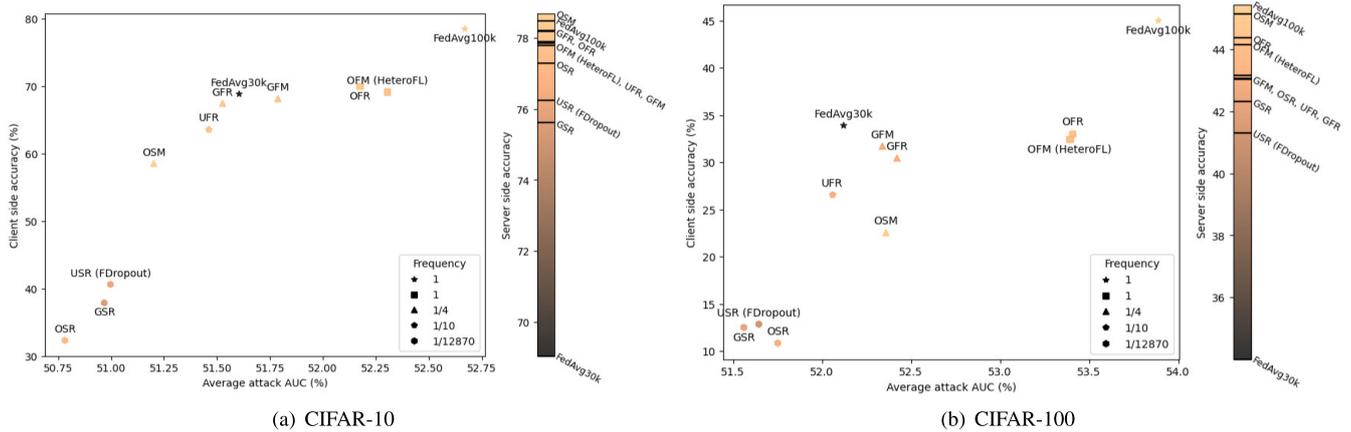

**FIGURE 2.** Client accuracy vs privacy trade-off and server-side accuracy of the 9 heterogeneous FL methods under study. Privacy attack performance (AUC) averaged over the 3 MIAs (Yeom, LiRA, tMIA). All heterogeneous FL architectures consist of 2 clients with large models (100k parameters, same size as the server's model) and 8 clients with small models (30k parameters). For reference, we report the performance of FedAvg30k and FedAvg100k with 10 small (30k parameters) and 10 large (100k parameters) clients, respectively. Heterogeneous FL methods with optimal client accuracy-privacy trade-off would be on the top-left corner of the graph. Heterogeneous FL methods with largest server-side accuracy are depicted at the top of the bar with gradient shading on the right-hand side of the graphs. Marker highlights repeated channel frequency.

### 3) MEMBERSHIP INFERENCE ATTACKS
Each client is subject to the 3 previously described MIAs. For LiRA and tMIA, the auxiliary dataset is drawn from the datasets of the rest of the clients $\mathbb{D}_a^c = \{\mathbb{D}_1, \ldots, \mathbb{D}_C\} \setminus \mathbb{D}_c$. We use the same shadow models to attack models from the same experiment. We train 16 shadow models for LiRA and use 25 distillation epochs for tMIA.

### 4) PERFORMANCE METRICS
We report three performance metrics: (1) client-side and (2) server-side accuracies on the testsets; and (3) the average AUC of the 3 MIAs. We select AUC because it is the metric that exhibited the largest correlation across MIAs on the evaluation datasets. Table 2 depicts the correlation of the three performance metrics (AUC, attack advantage and TPR@FPR 0.1) of the three MIAs (tMIA, LiRA, Yeom) on the CIFAR-10 dataset for illustration.

### C. RESULTS
#### 1) PRIVACY – PERFORMANCE TRADE-OFF
Figure 2 depicts the three performance measures of study, namely client-side accuracy (Y-axis), average attack AUC of the 3 MIAs (X-axis) and server-side accuracy (bar with gradient shading), of the nine heterogeneous FL methods in a federation with 2 *large* clients on the CIFAR-10 (a) and CIFAR-100 (b) datasets. The complete set of results can be found in Tables 8 and 9 in the Appendix.

The results corroborate our first hypothesis *H1* related to the accuracy-privacy trade-off. From a client perspective, methods GFM and GFR achieve similar accuracies as HeteroFL but with with better privacy protection ($0.5 - 1.0\%$ AUC). Their overall accuracy-privacy trade-off is similar to that of FedAvg30k yet they achieve significantly better server-side accuracy (**77.89**% and **78.19**% over 69.04% for CIFAR-10 and **43.05**% and **43.17**% over 34.01% for CIFAR-100). FDropout, and the GSR and OSR methods perform well in terms of client privacy, but their client accuracy is significantly lower when compared to the rest of the methods.

Supporting our *H2* hypothesis, methods GFR and GFM, and methods OFR and OFM yield similar results in all three measures on the two datasets, with OFM (HeteroFL) and OFR on CIFAR-100 being the closest with differences of only 0.2%, 0.6%, and 0.02% on the server accuracy, client accuracy, and attack AUC, respectively.

Interestingly, while the OSM method performs as expected on the client side, it outperforms every other FL method on its server-side accuracy, providing the best server accuracy-client privacy trade-off from all the studied methods.

#### 2) IMPACT OF THE NUMBER OF CLIENTS WITH SMALL MODEL COMPLEXITY
To evaluate hypothesis *H3*, we perform experiments with heterogeneous federations with 10 clients of which 2, 5 or 8 clients learn models of the same model complexity as the server's model. Table 3 summarizes the difference in performance between the best and the worse performing methods in each of these federations. Note how the difference in performance between the best and worst performing models in a federation with 2 *large* clients vs a federation with 8 *large* clients is **3x** for the server-side accuracy and attack AUC, and over **6x** for the client-side accuracy. These results support hypothesis *H3*.

#### 3) NON-IID DATA
We study the impact of non-IID (non independent and identically distributed) data using the Federated EMNIST

or FEMNIST [37] dataset, which is an image dataset of hand-written characters. We select this dataset because it is common in both the MIA and FL literature: while the original MNIST dataset is frequently used to evaluate MIAs, [28] its federated version (FEMNIST) is commonly used to evaluate FL methods [38], [39]. It consists of 62 classes with a long-tail data distribution. In its federated version, the images are distributed by the ID of the writer whose handwriting they are. Following the official sub-sampling method, we select 20% of the data, keeping only writers with at least 300 samples and splitting into train-test datasets where the test dataset corresponds to images by unseen. This results in approximately 165 writers in the train set. We distribute the data among 10 clients following the standard practice in the literature [38]. We do not apply data augmentation on this dataset.

**TABLE 3.** Absolute differences in performance between the best and worse performing methods in federations with 2, 5, and 8 *large* clients. As the ratio of clients with the same model size as the server increases, the differences in performance between the methods decreases corroborating our third hypothesis.

|  | CIFAR-10 | CIFAR-100 |
|---|---|---|
| Server Acc (2 clients) | **Δ3.09** (75.61-78.70) | **Δ3.82** (41.31-45.13) |
| Server Acc (5 clients) | Δ0.56 (78.44-79.00) | Δ1.13 (44.57-45.70) |
| Server Acc (8 clients) | Δ0.72 (78.38-79.10) | Δ1.69 (44.85-46.54) |
| Client Acc (2 clients) | **Δ37.65** (32.30-69.95) | **Δ22.18** (10.85-33.02) |
| Client Acc (5 clients) | Δ20.61 (51.41-72.02) | Δ11.88 (24.27-36.08) |
| Client Acc (8 clients) | Δ6.63 (67.32-73.96) | Δ3.25 (37.03-40.06) |
| Attack AUC (2 clients) | **Δ1.52** (50.78-52.30) | **Δ1.85** (51.56-53.41) |
| Attack AUC (5 clients) | Δ0.87 (51.69-52.56) | Δ1.32 (52.61-53.85) |
| Attack AUC (8 clients) | Δ0.65 (52.40-53.06) | Δ0.99 (52.91-53.90) |

While the previously formulated hypotheses hold in the case of non-IID data on the client-side, the server-side accuracy significantly drops when using heterogeneous FL methods: 2.2 points for FL with 2 *large* clients, and 0.9 points for 5 *large clients*. Furthermore, the `FedAvg100k` baseline outperforms several of the studied methods regarding client privacy. These results shed light on the limitations of the studied model integration methods in heterogeneous FL, and suggest that further research is needed to develop novel heterogeneous FL methods that consider the spurious correlations within the clients [40]. The Appendix contains more details about the experiments with non-IID data, including the results of applying two popular approaches to mitigate the challenges associated with non-IID data. The results, summarized in Table 6, illustrate how most methods improve both in accuracy and privacy protection yet there is no single method that yields the best overall performance, highlighting the importance of considering the model integration strategy in heterogeneous FL settings.

## VI. CONCLUSION AND FUTURE WORK

In this paper, we have proposed a novel taxonomy of heterogeneous FL methods that not only frames existing approaches into the same family of methods, but also enables the proposal of 7 new methods. In extensive empirical evaluations with the CIFAR-10 and CIFAR-100 datasets, we have studied the server-side accuracy and the client accuracy-privacy trade-off of these approaches when subjected to three commonly used membership inference attacks. Our results show that heterogeneous FL models can be used to mitigate the vulnerability against such attacks. Moreover, the strategy adopted to integrate the clients' models into the server's model impacts both the accuracy and privacy of the federation. By establishing a comprehensive taxonomy and introducing novel methodologies, we pave the way for enhanced privacy of sensitive data within federated learning environments. In future work, we plan to develop more robust methods to consider non-IID data in the clients in heterogeneous FL settings, particularly when there might be spuriou correlations present in the datasets.

## APPENDIX A
## MACHINE LEARNING MODEL

The model in all our experiments is a Convolutional Neural Network, similar to the models reported in related work [5]. The layers with weight matrices consist of 2D convolutional layers with a $(N, M, H, W)$ 4-dimensional matrix, where the first two dimensions $N$ and $M$ correspond the output and input channels and the rest are the convolutional kernels. From a heterogeneous FL perspective, $N$ and $M$ are the dimensions that change when the clients in the federation learn models of different size than the server's model whereas $H$ and $W$ are the same as in the server. In the PyTorch implementation, the bias of the convolutional layers has a separate $(N)$ 1-dimensional matrix. When a subset of $N_c^l$ output channels is selected for a client $c$ and convolutional layer $l$, its bias shares the same $N_c^l$ out of $N^l$ output channels.

After the convolutional layers in the model architecture, there are BatchNorm normalization layers with $(N)$ 1-dimensional weight matrices with bias. Note that the BatchNorm layer $l+2$ after convolutional layer $l$ has the same $N_c^{l+2} = N_c^l$ channels selected. The Scaler layer adapted from `HeteroFL` [5] scales its input with respect to the model-agnostic compression rate. For $r_c = \frac{N_c}{N} = \frac{M_c}{M}$, the Scaler follows:

$$f_{\text{Scaler}}(x) = \frac{1}{r_c}x. \quad (6)$$

Finally, there is a linear layer *lin* with weight matrix $(N, M)$ and bias with weight matrix size $(N)$. Each client $c$ shares the same $N_c^{lin}$ output channels in this linear layer.

The complexity of the model is controlled with parameter $u$. Each input and output dimension of the weight matrix is a multiple of $u$. The model complexity levels used in this paper –namely 30k, 100k, 400k, and 1.6M– correspond to $u$ values of 8, 16, 32, and 64, respectively. Figure 3 illustrates the model architecture for a generic $u$ and an example with $u = 16$.

| Fix model archtecture | | Changing size | | | |
|---|---|---|---|---|---|
| | | u=16 | | u | |
| Input (32x32x3) | (HxW) | N | M | N | M |
| Conv2D | 3x3 | 16 | 3 | u | 3 |
| Scaler | | | | | |
| BatchNorm | | 16 | | u | |
| ReLU | | | | | |
| MaxPool2D | | | | | |
| Conv2D | 3x3 | 32 | 16 | 2u | u |
| Scaler | | | | | |
| BatchNorm | | 32 | | 2u | |
| ReLU | | | | | |
| MaxPool2D | | | | | |
| Conv2D | 3x3 | 64 | 32 | 4u | 2u |
| Scaler | | | | | |
| BatchNorm | | 64 | | 4u | |
| ReLU | | | | | |
| MaxPool2D | | | | | |
| Conv2D | 3x3 | 128 | 64 | 8u | 4u |
| Scaler | | | | | |
| BatchNorm | | 128 | | 8u | |
| ReLU | | | | | |
| GlobalAveragePool2D | | | | | |
| Flatten | | | | | |
| Dense | | 10 | 128 | 10 | 8u |
| Output (10) | | | | | |

**FIGURE 3.** Model architecture and sizes of the weight matrices depending on the model complexity, controlled by the parameter $u$. Layer names and constant parameter dimensions on the left, varying dimensions on the right.

## APPENDIX B
## MEMBERSHIP INFERENCE ATTACKS
### A. YEOM ATTACK

The Yeom attack [28] is a membership inference attack that relies on comparing the prediction loss of a model on specific data instances to a pre-defined threshold. This threshold distinguishes between instances that were likely part of the training dataset and those that were not. The underlying assumption is that data points used in training tend to have a lower loss than those that were not, because the model has learned to perform well on training instances.

### 1) ATTACK SETUP AND THRESHOLD SELECTION
The `Yeom` attack relies on two main components:
1) **Global Threshold** ($v$): The attacker sets a loss threshold $v$, below which an instance is considered likely to belong to the training dataset.
2) **Attacker's Knowledge**: To calculate this threshold, the attacker uses a subset of data instances with known membership. This auxiliary set includes:
   - $\mathbb{D}_{\mathcal{A}+}$: Known training instances (member data).
   - $\mathbb{D}_{\mathcal{A}-}$: Known non-training instances (non-member data).

Using $\mathbb{D}_{\mathcal{A}+}$, the attacker computes the threshold $v$ as the average loss on known member instances:

$$v = \frac{1}{|\mathbb{D}_{\mathcal{A}+}|} \sum_{(x',y') \in \mathbb{D}_{\mathcal{A}+}} \mathrm{loss}(y', f(x'; \theta))$$

where:
- $f(x'; \theta)$ is the model's prediction for input $x'$,
- $\mathrm{loss}(y', f(x'; \theta))$ is the loss for true label $y'$ and prediction $f(x'; \theta)$.

### 2) MEMBERSHIP INFERENCE DECISION
Once the threshold $v$ is established, the attacker determines the membership status of a new instance $(x, y)$ by comparing its loss to $v$:

$$\mathcal{A}_{\texttt{Yeom}}(\hat{y}, (x, y)) = \begin{cases} 1, & \text{if } \mathrm{loss}(y, \hat{y}) < v \\ 0, & \text{otherwise.} \end{cases}$$

Here, $\hat{y} = f(x; \theta)$ is the model's predicted label for $x$, and $\mathrm{loss}(y, \hat{y})$ is the computed loss for this instance.

### B. LiRA ATTACK
We use the offline version of the `LiRA` (Likelihood Ratio Attack) attack of [29].

### 1) ATTACK SETUP: SHADOW MODELS AND AUXILIARY DATASET
The attack operates in several steps:
1) **Auxiliary Dataset** ($\mathbb{D}_a$): The attacker has access to an auxiliary dataset $\mathbb{D}_a$ that is similar to the data used to train the target model, although not necessarily identical. This auxiliary data is used to simulate the behavior of the target model with respect to training and non-training instances.
2) **Shadow Models** ($M_{sw}$): Using $\mathbb{D}_a$, the attacker trains multiple "shadow models" $M_{sw}$ on different random subsets $\mathbb{D}_{sw} \subset \mathbb{D}_a$. Each shadow model is designed to mimic the target model's behavior, especially in terms of confidence levels for data that was and was not in the training set.
3) **Confidence Scores**: For each data instance $(x, y)$, the attacker queries each shadow model $M_{sw}$ to obtain a confidence score, typically the model's probability prediction for the true label $y$. We denote the confidence score of shadow model $M_{sw}$ on $(x, y)$ as $\phi(M_{sw}(x), y)$.

### 2) MODELING CONFIDENCE SCORES WITH A GAUSSIAN DISTRIBUTION
For the data instance $(x, y)$, the attacker gathers confidence scores from all shadow models, yielding the set $\{\phi(M_1(x), y), \phi(M_2(x), y), \ldots, \phi(M_k(x), y)\}$. This set of scores is used to model the distribution of confidence values for instances that are either "in" or "out" of the training data.

The attacker then fits a Gaussian distribution $\mathcal{N}(\mu, \sigma^2)$ to these confidence scores, where $\mu$ and $\sigma^2$ represent the mean and variance of the scores. This Gaussian distribution captures the typical confidence score behavior of shadow models, depending on whether $(x, y)$ was in the training data.

#### 3) CALCULATING MEMBERSHIP PROBABILITY

Once the Gaussian distribution $\mathcal{N}(\mu, \sigma^2)$ is fitted, the attacker uses it to assess the probability that a new confidence score $\phi(M(x), y)$ from the target model $M$ is characteristic of training data.

The membership probability is calculated as:

$$1 - \Pr\left[\mathcal{N}(\mu, \sigma^2) > \phi(M(x), y)\right]$$

This probability reflects the likelihood that $\phi(M(x), y)$ would be a typical score under the Gaussian model. A higher probability suggests that $(x, y)$ is likely part of the training data, whereas a lower probability indicates non-membership.

#### 4) MEMBERSHIP INFERENCE DECISION

To make a final membership inference decision, the `LiRA` attack applies a threshold $\nu$ to determine whether the probability of membership exceeds a certain level as per the following decision rule:

$$\mathcal{A}_{\text{LiRA}}(\hat{y}, (x, y), \mathbb{D}_a) = \begin{cases} 1, & \text{if } 1 - \Pr[\mathcal{N}(\mu, \sigma^2) > \\ & \phi(\hat{y}, y)] < \nu \\ 0, & \text{otherwise.} \end{cases}$$
(7)

such that
- The attack outputs 1 (indicating membership) if the membership probability exceeds the threshold $1 - \nu$.
- Otherwise, it outputs 0 (indicating non-membership).

### C. TRAJECTORY MIA ATTACK

The `tMIA` attack [30] determines the membership of a data point based on the loss trajectory of the instance over multiple training epochs. The underlying hypothesis is that the sequence of loss values (i.e., the *loss trajectory*) of a data point changes differently for training data (members) and non-training data (non-members) as the model learns. Therefore, tracking these loss values over epochs can reveal membership status.

In a black-box setting such as ours, only the final trained model is accessible, and therefore the loss trajectory throughout training is not available. To address this, the `tMIA` attack uses knowledge distillation to approximate the loss trajectory.

#### 1) ATTACK SETUP: TARGET AND SHADOW MODELS

The target model is denoted by $M_{tg}^0(f, \mathbb{D}_g)$, where $f$ is the model function and $\mathbb{D}_g$ is the dataset used to train the model. The attack involves two key steps:
1) **Shadow Model Training**: A shadow model $M_{sw}^0$-$(f, \mathbb{D}_{sw}^+)$ is trained on a subset $(\mathbb{D}_{sw}^+, \mathbb{D}_{sw}^-) \subset \mathbb{D}_a$, where $\mathbb{D}_{sw}^+$ contains samples similar to the training data and $\mathbb{D}_{sw}^-$ contains non-training samples.
2) **Distillation Process**: The attacker distills both the target and shadow models on a distillation dataset $\mathbb{D}_{dl} \subset \mathbb{D}_a$. During this process, snapshots of the models are saved at each training epoch, resulting in a sequence of models:

$$\{M_{tg}^0, M_{tg}^1, \ldots, M_{tg}^d\} \quad \text{and} \quad \{M_{sw}^0, M_{sw}^1, \ldots, M_{sw}^d\}.$$

#### 2) LOSS TRAJECTORY CALCULATION

For a data instance $(x, y)$, its loss trajectory is captured by evaluating the loss of the data point at each epoch during the distillation process. This yields a sequence of losses:

$$\lambda_*^{(x,y)} = \{l_*^0, l_*^1, \ldots, l_*^d\}_*^{(x,y)}$$

where each $l_*^i$ is the loss at epoch $i$ for model $M_*^i$, and $* \in \{tg, sw\}$ represents either the target or shadow model.

#### 3) TRAINING THE ATTACK MODEL

The attack model $M_A$ is trained to recognize patterns in the loss trajectories that indicate membership. To train $M_A$, the loss trajectories of data points in the shadow model, both from the shadow training set $\mathbb{D}_{sw}^+$ (members) and shadow non-training set $\mathbb{D}_{sw}^-$ (non-members), are used. Specifically, the training set for $M_A$ consists of:

$$\{\lambda_{sw}^{(x,y)} \mid (x, y) \in \mathbb{D}_{sw}^+ \cup \mathbb{D}_{sw}^-\}$$

where each $\lambda_{sw}^{(x,y)}$ is the loss trajectory of $(x, y)$ on the shadow model.

#### 4) MEMBERSHIP INFERENCE DECISION

During inference, the attack model $M_A$ predicts the membership status of a new data instance $(x, y)$ by analyzing the loss trajectory from the target model's distillation process, $\lambda_{tg}^{(x,y)}$. The attack decision rule is:

$$\mathcal{A}_{\text{tMIA}}(\hat{y}, (x, y), \mathbb{D}_a) = \begin{cases} 1, & \text{if } M_A(\lambda_{tg}^{(x,y)}) > \nu \\ 0, & \text{otherwise.} \end{cases}$$
(8)

Here:
- $M_A(\lambda_{tg}^{(x,y)})$ is the output of the attack model on the loss trajectory $\lambda_{tg}^{(x,y)}$.
- $\nu$ is a threshold value; if the attack model's output exceeds this threshold, the instance is predicted as a member (1), otherwise, it is predicted as a non-member (0).

### APPENDIX C
### INPUT-OUTPUT CHANNEL DEPENDENCY

In section III-C, we present `FDropout` [4] and `HeteroFL` [5] according to their original descriptions, which suggest that the channels of a layer $l$ can be dropped independently from the previous and following channels. However, after extensive experiments, we observed that the client models train significantly better if the selected output channels of a convolutional layer are *the same* as the input channels of the following convolutional layer. In the `FDropout` adaptation of [7] the same principled is adopted: layer $l$ only drops output channels randomly, while the selection of the input channels is inherited from the previous convolutional layer. The pseudo-code in [41] suggests that their implementation follows the original layer-independent dropout and their results show that `FDropout` performs badly compared to other techniques: while the Simple Ensemble Averaging method reached the 70% accuracy of the baseline `FedAvg` on FEMNIST dataset, the presented implementation of `FDropout` only reached 60%. In table 4, we compare `FDropout` (USR) and GFR with input and output channels dropped independently and with layer-wise coupling with respect to the previous and following layer. The results show that the client side accuracy for the laye-wise methods outperforms their independent counterpart by 16% for `FDropout` and 7% for GFR. Based on these results we conclude that the layer-wise dependency is necessary to achieve competitive results and follow this principle in our other experiments.

Additionally, in this Appendix we describe how the BatchNorm layers in our implementation have the same channels dropped as the previous convolutional layers.

**TABLE 4.** Input-output channels selected independently and with respect to the previous layers in the CNN. `FDropout` (USR) and GFR experiments on CIFAR-10 with 2 large clients out of 10 clients in total, repeated 3 times. Client-side performance is significantly better when the channel selection is structured layer-wise compared to their independent counterparts. Privacy evaluated with the `Yeom` attack.

| Name | Server | | Client average | |
|---|---|---|---|---|
| | Acc ↑ | Adv ↓ | Acc ↑ | Adv ↓ |
| USR independent | 75.44 ± 1.75 | 2.57 ± 1.62 | 23.56 ± 0.29 | 1.74 ± 0.44 |
| USR layerwise | **76.38** ± 1.36 | **2.45** ± 1.34 | **39.99** ± 1.11 | 2.32 ± 1.52 |
| GFR indepentent | 76.41 ± 1.73 | 2.87 ± 0.78 | 55.82 ± 1.61 | 3.00 ± 0.50 |
| GFR layerwise | **77.11** ± 1.52 | 3.34 ± 0.98 | **62.80** ± 1.43 | 3.20 ± 0.24 |

## APPENDIX D
## DATASET SIZE, PRIVACY, MODEL SIZE AND ACCURACY

Previous work has shown that as models get more complex, they are more vulnerable to MIAs. For example, [28] demonstrate that their attack's accuracy increases as the model size increases on standard benchmark image datasets. In FL, Li et al. [9] reported that, the larger the models, the more vulnerable they are to model memorization attacks. In their case, it was a horizontal FL architecture with the same model (ResNet) both in the server and the clients. Other works have highlighted that over-parameterized models are vulnerable to membership memorization attacks [42].

In this section, we shed further light on this topic by focusing on the privacy-accuracy trade-off in FL with respect to dataset and model size, and from the perspective of both the server and the clients. Note that prior studies have only analyzed the server's performance. By means of an empirical illustrative example, we show that, for a given model and an FL scenario, there is a strong negative correlation between the size of the clients' datasets and models, and their vulnerability against membership inference attacks (`Yeom` in our example). As previously discussed, this attack occurs on the last update the client sends to the server in round $T$, $\mathcal{A}_{\texttt{Yeom}}(\boldsymbol{\theta}_c^T)$. We use the `Yeom` attack for this illustrative experiment as it requires significantly less computation than the other described MIAs.

We perform the experiments on the CIFAR-10 image dataset (see Section V for a description of the dataset) with 10 homogenous clients and a `FedAvg` FL architecture [1]. In `FedAvg` the clients train the same model as the server using their own dataset, such that the average of the clients' model weights is an approximation of training the same model in a centralized machine with access to all client data. That is, `FedAvg` computes $\min_\theta L(\boldsymbol{\theta})$, given by: $\min_\theta L(\boldsymbol{\theta}) =$

$$\min_{\boldsymbol{\theta}} \frac{1}{|\mathbb{D}|} \sum_{c=1}^{C} \sum_{(\boldsymbol{x},y)\in\mathbb{D}_c} l(y, f(\boldsymbol{x},\boldsymbol{\theta})) \approx \frac{1}{C}\sum_{c=1}^{C} \min_{\boldsymbol{\theta}_c} L_c(\boldsymbol{\theta}_c, \mathbb{D}_c)$$

(9)

where $L$ is the loss function in the server when having access to all the client data; $l$ is the loss function in each client; $\boldsymbol{\theta}$ and $f$ are server model parameters and server architecture, respectively. The loss at each client $L_c(\boldsymbol{\theta}_c, \mathbb{D}_c)$ is given by $\frac{1}{|\mathbb{D}_c|} \sum_{(\boldsymbol{x},y)\in\mathbb{D}_c} l(y, f(\boldsymbol{x}, \boldsymbol{\theta}_c))$, where $C$ is the number of clients; and $\mathbb{D}_c$ represents the dataset of client $c$ such that $\mathbb{D} = \bigcup_{c=1}^{C} \mathbb{D}_c$ corresponds to the entire dataset.

To ensure a fair evaluation, the attacker's knowledge dataset $\mathbb{D}_{\mathcal{A}+}$ for the `Yeom`'s attack is proportionate to the size of the training dataset. Specifically, we select 1%: $|\mathbb{D}_{\mathcal{A}+}| = \min(3, 0.01|\mathbb{D}_c|)$ for the attack on client $c$ with dataset size $|\mathbb{D}_c|$. The attack test dataset $\mathbb{D}_{\text{MIA}}$ contains the same number of samples from the training set as samples from outside of the training set. If the client $c$ has less than 5,000 data samples, we test on all of the client data samples with non-member examples from the test set, so that $|\mathbb{D}_{\text{MIA}}| = 2\mathbb{D}_c$, otherwise it is capped at 5,000. With such a dataset setting, a simple baseline which guesses that each MIA test data point is part of the training dataset would give a 50% accuracy. We define the attack advantage [43] as the improvement of an attack when compared to this baseline according to: $Adv(\mathcal{A}) = 2(Acc(\mathcal{A}) - 50)$, where $Acc(\mathcal{A})$ is the accuracy of the attacker's model.

Regarding the machine learning models, we adopt the architecture proposed in [5]. It consists of a convolutional neural network (CNN) with 4 convolutional layers and one fully connected layer at the end. We adjust the model complexity by changing the number of channels in the convolutional layers and the number of units in the last fully connected layer. We define 4 levels of model complexity and train 5 models for each level of complexity using `FedAvg` with class-balanced data in each client, resulting in 50 client models. The complexity of the models is measured by the number of parameters, ranging from models with 30k to models with 1.6 million parameters.

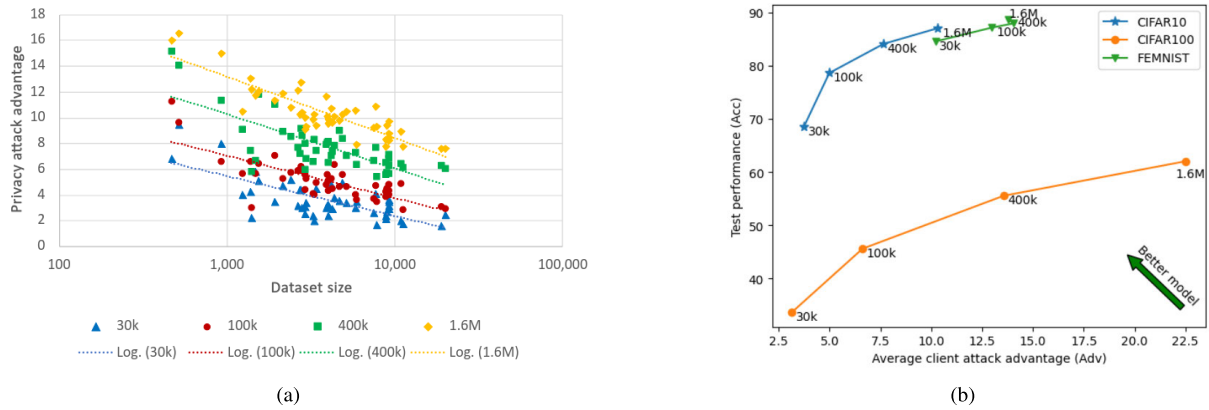

**FIGURE 4.** (a): Exemplary illustration of the correlation between the privacy attack advantage for the `Yeom` attack and the dataset size from the clients' perspective. Results for 5 repeated experiments on the CIFAR-10 dataset using the `FedAvg` architecture with 10 clients having different dataset sizes, resulting in 50 client models. Each dot depicts a client in one federated training and the color represents different model complexities (CNNs), characterized by the number of parameters, ranging from 30k to 1.6 million. Note the negative correlations between the size of the clients' dataset and the attack advantage, as well as between the model's complexity and the associated attack advantage. (b): Privacy-accuracy trade-off of the data depicted in (a) by averaging experiments across clients per model complexity. In addition to CIFAR-10, we also show the trade-off for the CIFAR-100 and FEMNIST datasets. The attacker's advantage and test accuracy on the clients increases as the model size increases. Observations in (a) and (b) suggest that model-agnostic Federated Learning could be a privacy-enhancing solution.

For each model complexity, we compute the Pearson correlation coefficient between the logarithm of the clients' dataset size, $\log_{10}(|\mathbb{D}_c|)$, and the attack advantage on the clients' final update, $Adv(\mathcal{A}_{\texttt{Yeom}}(\theta_c^T))$. Figure 4(a) visually illustrates the correlation between the client's dataset size and the attack advantage on the models of increasing complexity on the CIFAR-10 dataset. Note that clients with less than 400 data points are not considered in the calculation as their attack performance is not consistent through runs due to having very small (< 4) attacker knowledge. Figure 4(b) depicts the privacy-accuracy trade-off by averaging experiments across clients for each model complexity on the CIFAR-10, CIFAR-100, and FEMNIST datasets. We observe strong negative correlations between the size of the clients' dataset and the attack's advantage; and between the clients' model complexity and the corresponding attack's advantage. We also observe that both the attacker's advantage and the test accuracy on the clients increase as the model size increases. These results suggest that model-agnostic FL could enhance privacy both in the server and the clients by means of learning models in the clients that are smaller than the server's model.

## APPENDIX E
## DETAILED EXPERIMENTAL RESULTS
### A. MODEL SIZE VS ATTACK ADVANTAGE
Table 5 shows the Pearson correlation coefficient between the client dataset sizes and their vulnerability against client-side `Yeom` membership inference attacks. Numbers correspond to running experiments 5 times in a federation with 10 clients. Clients with less than 400 data samples are excluded from the analysis, resulting in the exclusion of 3 clients in the 5 runs with the CIFAR-10 and CIFAR-100 datasets. All values in the table exceed the critical value of non-significant correlation for the given sample size.

**TABLE 5.** Pearson correlation coefficient of dataset size and attack advantage for different model sizes in class-balanced heterogeneous data distribution.

| # Model parameters | 30k | 100k | 400k | 1.6M |
|---|---|---|---|---|
| CIFAR-10 | -0.62 | -0.65 | -0.57 | -0.87 |
| CIFAR-100 | -0.54 | -0.70 | -0.85 | -0.90 |
| FEMNIST | -0.71 | -0.65 | -0.63 | -0.75 |

**TABLE 6.** Performance and change in performance when FedProx or FedAvgM are applied to the Federated Learning methods. Improvements are highlighted in bold and the best performing results are underlined. As seen in the Table, most methods improve both in accuracy and privacy protection yet there is no single method that yields the best performance. The table illustrates the impact of the server model integration strategy in heterogeneous FL settings. As observed with IID data, randomness in the channel selection yields better privacy protection and competitive server accuracy at the expense of client accuracy.

| FL Method | FedProx($\mu = 0.01$) | | | FedAvgM($\eta = 0.9, \beta_1 = 0.3$) | | |
|---|---|---|---|---|---|---|
| | ↑ Server Acc | ↑ Client Acc | ↓ MIA (Yeom) AUC | ↑ Server Acc | ↑ Client Acc | ↓ MIA (Yeom) AUC |
| FedAvg100k | 87.37 (**0.17**) | 87.46 (**0.28**) | 56.19 (0.21) | 87.53 (**0.33**) | 87.50 (**0.32**) | 56.08 (0.10) |
| FedAvg30k | 84.40 (**0.30**) | 84.27 (**0.07**) | 55.13 (**-0.28**) | 84.22 (**0.12**) | 84.20 (**0.01**) | 54.91 (**-0.51**) |
| OFM (HeteroFL) | 83.09 (-0.10) | 82.82 (**0.23**) | 56.68 (**-0.06**) | 82.97 (-0.22) | 82.61 (**0.02**) | 56.60 (**-0.14**) |
| OFR | 82.49 (-0.46) | 83.08 (**0.16**) | 56.92 (0.43) | 83.08 (**0.14**) | 82.59 (-0.33) | 56.26 (**-0.23**) |
| OSM | 84.28 (-0.12) | 73.68 (**0.50**) | 55.77 (**-0.56**) | 84.33 (-0.07) | 73.95 (**0.77**) | 56.18 (**-0.15**) |
| GFM | 85.40 (**0.17**) | 81.82 (**0.01**) | 55.89 (**-0.29**) | 85.33 (**0.09**) | 81.82 (**0.02**) | 55.86 (**-0.32**) |
| GFR | 83.92 (**0.15**) | 81.59 (**0.26**) | 55.89 (**-0.21**) | 84.04 (**0.28**) | 82.01 (**0.69**) | 55.84 (**-0.26**) |
| UFR | 84.67 (**0.24**) | 79.58 (**0.46**) | 55.45 (**-0.09**) | 85.11 (**0.69**) | 79.05 (-0.07) | 55.55 (0.01) |
| OSR | 83.98 (**0.72**) | 32.07 (**0.91**) | 54.90 (**-0.24**) | 83.71 (**0.46**) | 28.93 (-2.24) | 55.52 (0.38) |
| GSR | 84.79 (**0.01**) | 44.75 (**3.16**) | <u>54.96</u> (**0.01**) | 84.39 (-0.39) | 44.34 (**2.75**) | 55.21 (0.26) |
| USR (FDropout) | <u>85.62</u> (**0.19**) | 46.90 (**2.85**) | 54.98 (0.03) | <u>85.29</u> (-0.14) | 45.03 (**0.98**) | <u>54.98</u> (**0.02**) |

### B. EXPERIMENTS ON NON-IID DATA
Figure 5 depicts the performance of the 9 model-agnostic FL methods on the FEMNIST dataset for a federation with 2 and 5 *large* (100k parameters) clients and the same experimental setup as that described in the main paper.

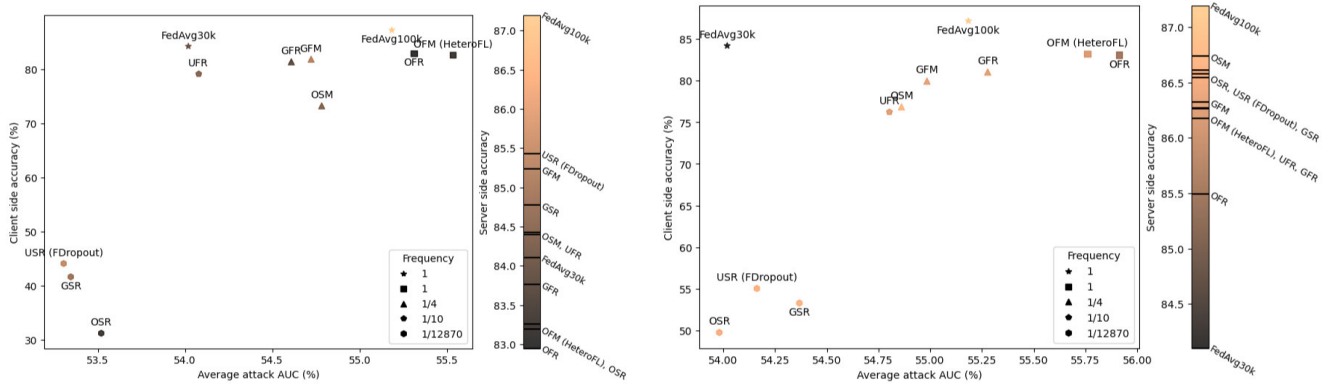

(a) Performance of the 9 model-agnostic FL methods and baselines on the FEMNIST dataset with 2 *large* (100k parameters) clients. The server-side accuracy of several model-agnostic FL methods is inferior to `FedAvg30k`'s accuracy. Contrary to the results obtained on the CIFAR-x datasets, `FDropout` is the model-agnostic FL method with the best server-side accuracy.

(b) Performance of the 9 model-agnostic FL methods and baselines on the FEMNIST with 5 *large* (100k parameters) and 5 *small* (30k parameters) clients. The non-IID dataset requires more *large* clients in the federation to outperform the `FedAvg30k` baseline on the server-side accuracy. The most privacy-sensitive model-agnostic approaches, such as `OFM (HeteroFL)` yield worse client accuracy-privacy trade-off than the `FedAvg100k` baseline in this non-IID setting.

**FIGURE 5.** Performance of the 9 model-agnostic methods and baselines on the FEMNIST dataset with 2 and 5 *large* (100k parameters) clients. These results suggest that more sophisticated model-agnostic approaches that take into account spurious correlations beyond channel selection strategies are needed.

**TABLE 7.** Detailed results on the FEMNIST dataset. Experiments averaged over 3 runs. Best in each category highlighted with bold. Methods are grouped by number of *large* clients and ordered based on the frequency a client receives the same parameters.

| # large clients | Freq. | Method name | ↑ Server Acc % | ↑ Client Acc % | ↓ AUC % | | | | ↓ Attack Adv % | | | | ↓ TPR% at 0.1% FPR | | | |
|---|---|---|---|---|---|---|---|---|---|---|---|---|---|---|---|---|
| | | | | | tMIA | LiRA | Yeom | Avg | tMIA | LiRA | Yeom | Avg | tMIA | LiRA | Yeom | Avg |
| 10 | 1 | FedAvg100k | 87.20 ± 0.46 | 87.18 ± 0.42 | 56.41 ± 0.43 | 53.17 ± 0.25 | 55.98 ± 0.10 | 55.18 ± 0.10 | 7.67 ± 0.23 | 5.20 ± 0.38 | 7.26 ± 0.42 | 6.71 ± 0.25 | 0.27 ± 0.02 | 0.25 ± 0.06 | 0.15 ± 0.03 | 0.22 ± 0.01 |
| 0 | 1 | FedAvg30k | 84.10 ± 0.97 | 84.20 ± 1.04 | 55.45 ± 0.51 | 51.20 ± 0.16 | 55.41 ± 0.58 | 54.02 ± 0.38 | 6.86 ± 0.71 | 2.26 ± 0.26 | 7.01 ± 0.61 | 5.38 ± 0.38 | 0.15 ± 0.02 | 0.14 ± 0.02 | 0.16 ± 0.02 | 0.15 ± 0.01 |
| 2 | 1 | OFM (HeteroFL) | 83.19 ± 0.83 | 82.59 ± 0.26 | 57.24 ± 0.73 | 52.62 ± 0.24 | 56.74 ± 0.37 | 55.53 ± 0.29 | 9.21 ± 0.43 | 4.02 ± 0.51 | 8.50 ± 0.54 | 7.24 ± 0.40 | 0.24 ± 0.04 | 0.16 ± 0.05 | 0.13 ± 0.03 | 0.18 ± 0.01 |
| 2 | 1 | OFR | 82.95 ± 0.09 | 82.92 ± 1.04 | 56.93 ± 0.63 | 52.51 ± 0.34 | 56.49 ± 0.60 | 55.31 ± 0.35 | 8.68 ± 0.92 | 3.09 ± 1.07 | 8.17 ± 1.00 | 6.65 ± 0.83 | 0.28 ± 0.06 | 0.11 ± 0.03 | 0.11 ± 0.01 | 0.17 ± 0.02 |
| 2 | 1/4 | OSM | 84.39 ± 0.60 | 73.18 ± 0.81 | 56.47 ± 0.46 | 51.55 ± 0.39 | 56.33 ± 0.48 | 54.78 ± 0.18 | 8.60 ± 0.57 | 2.35 ± 0.64 | 8.39 ± 0.72 | 6.44 ± 0.26 | 0.22 ± 0.03 | 0.15 ± 0.06 | **0.08** ± 0.03 | 0.15 ± 0.02 |
| 2 | 1/4 | GFM | 85.23 ± 0.21 | 81.80 ± 0.19 | 56.27 ± 0.58 | 51.72 ± 0.17 | 56.18 ± 0.56 | 54.72 ± 0.30 | 8.29 ± 0.99 | 2.69 ± 0.18 | 7.66 ± 1.11 | 6.21 ± 0.45 | 0.22 ± 0.02 | 0.16 ± 0.02 | 0.13 ± 0.02 | 0.17 ± 0.01 |
| 2 | 1/4 | GFR | 83.76 ± 0.71 | 81.32 ± 0.27 | 55.95 ± 0.84 | 51.78 ± 0.18 | 56.10 ± 0.62 | 54.61 ± 0.41 | 7.56 ± 0.82 | 2.39 ± 0.88 | 7.21 ± 1.41 | 5.72 ± 0.46 | 0.20 ± 0.13 | 0.15 ± 0.05 | 0.18 ± 0.07 |  |
| 2 | 1/10 | UFR | 84.42 ± 0.39 | 79.12 ± 0.01 | 55.53 ± 0.27 | 51.15 ± 0.37 | 55.54 ± 0.34 | 54.08 ± 0.12 | 7.13 ± 0.80 | 1.95 ± 0.97 | 6.44 ± 0.38 | 5.17 ± 0.23 | **0.16** ± 0.03 | 0.13 ± 0.05 | 0.13 ± 0.03 | 0.14 ± 0.00 |
| 2 | $1/10^4$ | OSR | 83.26 ± 1.22 | 31.17 ± 1.30 | 54.93 ± 0.59 | 50.49 ± 0.46 | 55.14 ± 0.77 | 53.52 ± 0.43 | 6.26 ± 0.88 | 1.06 ± 0.89 | 6.02 ± 0.86 | 4.45 ± 0.53 | 0.17 ± 0.04 | 0.12 ± 0.03 | 0.09 ± 0.04 | 0.13 ± 0.02 |
| 2 | $1/10^4$ | GSR | 84.78 ± 1.01 | 41.59 ± 1.78 | 54.80 ± 1.25 | 50.28 ± 0.57 | **54.95** ± 0.34 | 53.35 ± 0.07 | **5.89** ± 0.11 | 0.73 ± 1.22 | 5.67 ± 1.01 | **4.10** ± 0.72 | **0.16** ± 0.01 | **0.07** ± 0.01 | 0.10 ± 0.02 | **0.11** ± 0.02 |
| 2 | $1/10^4$ | USR (FDropout) | **85.43** ± 0.54 | 44.05 ± 0.93 | **54.78** ± 0.25 | **50.18** ± 0.37 | **54.95** ± 0.47 | **53.31** ± 0.22 | 6.20 ± 1.02 | **0.69** ± 0.69 | **5.76** ± 0.89 | 4.22 ± 0.14 | 0.17 ± 0.05 | 0.10 ± 0.01 | 0.10 ± 0.02 | 0.12 ± 0.02 |
| 5 | 1 | OFM (HeteroFL) | 86.18 ± 0.25 | **83.21** ± 0.30 | 57.19 ± 0.00 | 52.86 ± 0.20 | 57.23 ± 0.14 | 55.76 ± 0.02 | 8.88 ± 0.41 | 4.46 ± 0.86 | 8.70 ± 0.11 | 7.35 ± 0.46 | 0.28 ± 0.01 | 0.19 ± 0.04 | **0.05** ± 0.02 | 0.17 ± 0.02 |
| 5 | 1 | OFR | 85.49 ± 0.24 | 83.10 ± 0.12 | 57.60 ± 0.33 | 52.99 ± 0.24 | 57.15 ± 0.25 | 55.91 ± 0.07 | 9.28 ± 0.70 | 5.24 ± 0.33 | 8.71 ± 1.50 | 7.74 ± 0.57 | 0.27 ± 0.03 | 0.20 ± 0.06 | 0.06 ± 0.01 | 0.18 ± 0.02 |
| 5 | 1 | OSM | **86.74** ± 0.47 | 76.85 ± 1.05 | 56.20 ± 0.08 | 52.16 ± 0.34 | 56.22 ± 0.55 | 54.86 ± 0.29 | 7.42 ± 0.49 | 2.98 ± 0.38 | 7.57 ± 1.02 | 5.99 ± 0.54 | **0.20** ± 0.06 | 0.16 ± 0.02 | 0.06 ± 0.01 | **0.14** ± 0.02 |
| 5 | 1/4 | GFM | 86.33 ± 0.19 | 79.94 ± 0.36 | 56.55 ± 0.42 | 52.03 ± 0.22 | 56.37 ± 0.19 | 54.98 ± 0.14 | 9.23 ± 0.02 | 2.86 ± 0.75 | 7.21 ± 0.16 | 6.44 ± 0.30 | 0.25 ± 0.05 | 0.13 ± 0.01 | **0.05** ± 0.03 | **0.14** ± 0.01 |
| 5 | 1/4 | GFR | 86.27 ± 0.07 | 81.04 ± 0.29 | 56.86 ± 0.17 | 52.33 ± 0.44 | 56.64 ± 0.50 | 55.28 ± 0.34 | 8.52 ± 1.24 | 3.65 ± 0.59 | 8.08 ± 0.87 | 6.75 ± 0.62 | 0.26 ± 0.05 | 0.15 ± 0.04 | 0.08 ± 0.01 | 0.16 ± 0.03 |
| 5 | 1/10 | UFR | 86.27 ± 0.54 | 76.24 ± 0.49 | 56.24 ± 0.23 | 51.99 ± 0.28 | 56.18 ± 0.40 | 54.80 ± 0.18 | 8.09 ± 0.23 | 3.11 ± 0.20 | 8.06 ± 0.92 | 6.42 ± 0.31 | 0.30 ± 0.09 | 0.12 ± 0.03 | 0.07 ± 0.03 | 0.16 ± 0.03 |
| 5 | $1/10^4$ | OSR | 86.54 ± 0.22 | 49.75 ± 3.31 | **55.35** ± 0.18 | **51.31** ± 0.45 | 55.28 ± 0.51 | **53.98** ± 0.35 | 6.91 ± 0.57 | **1.68** ± 0.24 | **6.33** ± 0.53 | **4.97** ± 0.36 | 0.23 ± 0.06 | **0.11** ± 0.04 | 0.07 ± 0.03 | **0.14** ± 0.03 |
| 5 | $1/10^4$ | GSR | 86.61 ± 0.07 | 53.28 ± 2.41 | 55.62 ± 0.14 | 51.64 ± 0.38 | 55.84 ± 0.22 | 54.37 ± 0.03 | 6.95 ± 1.57 | 2.36 ± 1.11 | 6.71 ± 0.85 | 5.34 ± 1.13 | 0.26 ± 0.02 | 0.14 ± 0.01 | 0.07 ± 0.05 | 0.16 ± 0.03 |
| 5 | $1/10^4$ | USR (FDropout) | 86.58 ± 0.53 | 55.04 ± 1.73 | 55.52 ± 0.41 | 51.38 ± 0.37 | 55.59 ± 0.49 | 54.16 ± 0.29 | **6.44** ± 0.88 | 2.10 ± 0.49 | 6.75 ± 0.84 | 5.10 ± 0.50 | 0.25 ± 0.12 | 0.17 ± 0.04 | **0.05** ± 0.02 | 0.16 ± 0.02 |
| 8 | 1 | OFM (HeteroFL) | 87.04 ± 0.21 | **83.93** ± 0.33 | 57.97 ± 0.32 | 53.41 ± 0.75 | 57.73 ± 0.15 | 56.37 ± 0.20 | 8.96 ± 0.77 | 4.22 ± 1.97 | 8.62 ± 0.12 | 7.27 ± 0.04 | **0.27** ± 0.03 | 0.22 ± 0.04 | **0.05** ± 0.06 | 0.18 ± 0.04 |
| 8 | 1 | OFR | 86.47 ± 0.99 | 83.44 ± 1.12 | 57.44 ± 0.05 | 53.23 ± 0.10 | 57.47 ± 0.01 | 56.05 ± 0.01 | 9.32 ± 0.71 | 4.79 ± 1.34 | 8.40 ± 0.09 | 7.51 ± 0.18 | 0.32 ± 0.02 | 0.25 ± 0.13 | 0.07 ± 0.06 | 0.21 ± 0.07 |
| 8 | 1/4 | OSM | 87.33 ± 0.14 | 79.80 ± 1.58 | 56.77 ± 0.53 | 52.66 ± 0.26 | 56.87 ± 0.01 | 55.43 ± 0.09 | 9.18 ± 0.69 | 3.84 ± 0.77 | 7.58 ± 0.03 | 6.87 ± 0.08 | 0.28 ± 0.04 | **0.13** ± 0.05 | 0.12 ± 0.11 | 0.18 ± 0.03 |
| 8 | 1/4 | GFM | 87.38 ± 0.04 | 80.89 ± 0.80 | **56.29** ± 0.27 | 52.72 ± 0.19 | 56.63 ± 0.26 | 55.21 ± 0.01 | 8.37 ± 0.17 | 4.38 ± 0.56 | 7.50 ± 1.56 | 6.75 ± 0.39 | **0.27** ± 0.01 | 0.20 ± 0.04 | 0.13 ± 0.05 | 0.20 ± 0.03 |
| 8 | 1/4 | GFR | 87.12 ± 0.38 | 80.74 ± 0.68 | 56.86 ± 0.39 | 53.06 ± 0.49 | 56.79 ± 0.12 | 55.57 ± 0.01 | 8.98 ± 0.83 | 4.73 ± 0.87 | 8.52 ± 0.34 | 7.41 ± 0.45 | 0.23 ± 0.02 | 0.20 ± 0.04 | 0.07 ± 0.05 | **0.17** ± 0.04 |
| 8 | 1/10 | UFR | 87.05 ± 1.28 | 76.97 ± 0.44 | 56.86 ± 0.70 | 52.91 ± 0.57 | 56.82 ± 0.20 | 55.53 ± 0.11 | 9.03 ± 0.69 | 4.32 ± 1.53 | 8.86 ± 0.53 | 7.40 ± 0.10 | 0.31 ± 0.06 | 0.20 ± 0.04 | 0.10 ± 0.03 | 0.20 ± 0.02 |
| 8 | $1/10^4$ | OSR | 87.27 ± 0.05 | 72.63 ± 0.09 | 56.33 ± 0.20 | 52.62 ± 0.25 | **56.45** ± 0.35 | **55.13** ± 0.36 | 8.39 ± 1.25 | 3.86 ± 1.05 | 7.20 ± 1.37 | 6.48 ± 1.22 | 0.30 ± 0.03 | 0.20 ± 0.04 | 0.16 ± 0.01 | 0.22 ± 0.06 |
| 8 | $1/10^4$ | GSR | **87.64** ± 0.05 | 73.19 ± 0.40 | 56.46 ± 0.32 | **52.61** ± 0.17 | 56.47 ± 0.14 | 55.18 ± 0.10 | 8.50 ± 0.78 | **3.71** ± 0.42 | **7.16** ± 0.51 | **6.45** ± 0.29 | 0.32 ± 0.02 | 0.17 ± 0.04 | 0.10 ± 0.02 | 0.20 ± 0.03 |
| 8 | $1/10^4$ | USR (FDropout) | 86.99 ± 0.05 | 72.76 ± 0.59 | 56.40 ± 0.13 | 52.63 ± 0.36 | 56.50 ± 0.37 | 55.18 ± 0.05 | **8.07** ± 0.05 | 4.63 ± 1.16 | 7.70 ± 0.80 | 6.80 ± 0.67 | 0.30 ± 0.00 | 0.15 ± 0.01 | 0.12 ± 0.05 | 0.19 ± 0.01 |

In the case of a federation with 2 large clients (Figure Figure 5(a)), we observe significant differences in the server's performance when compared to the results obtained with the CIFAR-10 and CIFAR-100 datasets. Contrary to the CIFAR-x datasets, methods `GFR`, `OSR`, `OFM`, and `OFR` underperform in terms of server-side accuracy when compared to the `FedAvg30k` baseline, while methods `OFR` and `OFM` underperform in terms of privacy compared to the`FedAvg100k` baseline. These results suggest that heterogeneous FL architectures where some of the clients learn smaller models can lead to higher privacy risks, highlighting the importance of analyzing the impact of model integration in those settings, as we do in this paper.

In a federation with 5 large clients (Figure Figure 5(b)), the server-side results are more similar to those obtained on the CIFAR-x datasets: method `OSM` yields the best server-side accuracy and all the methods outperform `FedAvg30k`. Interestingly and contrary to the behavior on the CIFAR-x datasets, `FDropout` is competitive with the other methods on its server-side accuracy, yet it yields poor client-side accuracy.

To mitigate the observed decrease in performance and privacy of the models in the case of non-IID data, we have

**TABLE 8.** Detailed results on the CIFAR-10 dataset. Experiments averaged over 3 runs. Best in each category highlighted with bold. Methods are grouped by number of *large* clients and ordered based on the frequency a client receives the same parameters.

| No. of large | Freq. | Method name | ↑ Server Acc % | ↑ Client Acc % | ↓ AUC % tMIA | LiRA | Yeom | Avg | ↓ Attack Adv % tMIA | LiRA | Yeom | Avg | ↓ TPR% at 0.1% FPR tMIA | LiRA | Yeom | Avg |
|---|---|---|---|---|---|---|---|---|---|---|---|---|---|---|---|---|
| 10 | 1 | FedAvg100k | **78.48** ± 0.23 | **78.47** ± 0.18 | 53.62 ± 0.15 | 51.77 ± 0.77 | 52.62 ± 0.55 | 52.67 ± 0.42 | 3.94 ± 0.40 | 2.49 ± 0.17 | 3.03 ± 0.65 | 3.16 ± 0.41 | 0.16 ± 0.05 | **0.08** ± 0.01 | 0.14 ± 0.04 | 0.13 ± 0.03 |
| 0 | 1 | FedAvg30k | 69.04 ± 0.24 | 68.82 ± 0.21 | **52.17** ± 0.35 | **50.83** ± 0.15 | **51.81** ± 0.23 | **51.60** ± 0.14 | **2.11** ± 1.01 | **0.69** ± 0.54 | **1.66** ± 0.05 | **1.49** ± 0.18 | **0.09** ± 0.02 | 0.09 ± 0.00 | **0.09** ± 0.02 | **0.09** ± 0.01 |
| 2 | 1 | OFM (HeteroFL) | 77.79 ± 1.53 | 69.17 ± 0.17 | 53.30 ± 0.31 | 51.13 ± 0.16 | 52.47 ± 0.27 | 52.30 ± 0.16 | 3.22 ± 1.14 | 1.46 ± 0.33 | 3.04 ± 0.76 | 2.57 ± 0.24 | 0.10 ± 0.02 | **0.06** ± 0.01 | 0.12 ± 0.04 | 0.10 ± 0.02 |
| 2 | 1 | OFR | 78.22 ± 1.53 | **69.95** ± 1.18 | 53.30 ± 0.55 | 50.94 ± 0.56 | 52.28 ± 0.44 | 52.17 ± 0.31 | 2.60 ± 1.09 | 0.68 ± 1.55 | 1.95 ± 0.57 | 1.74 ± 0.34 | 0.14 ± 0.04 | 0.08 ± 0.02 | 0.11 ± 0.05 | 0.11 ± 0.03 |
| 2 | 1/4 | OSM | **78.70** ± 1.28 | 58.52 ± 1.91 | 51.73 ± 1.19 | 50.48 ± 0.45 | 51.40 ± 1.03 | 51.20 ± 0.84 | 2.16 ± 0.41 | 1.19 ± 0.67 | 2.38 ± 0.21 | 1.91 ± 0.29 | 0.10 ± 0.01 | 0.08 ± 0.03 | **0.07** ± 0.03 | **0.08** ± 0.02 |
| 2 | 1/4 | GFM | 77.89 ± 1.09 | 68.13 ± 1.02 | 52.54 ± 0.27 | 50.84 ± 0.48 | 51.98 ± 0.39 | 51.79 ± 0.33 | 3.01 ± 0.71 | 0.67 ± 0.87 | 2.04 ± 0.33 | 1.91 ± 0.06 | 0.09 ± 0.05 | 0.10 ± 0.03 | 0.12 ± 0.03 | 0.10 ± 0.03 |
| 2 | 1/4 | GFR | 78.19 ± 1.32 | 67.44 ± 1.26 | 52.17 ± 0.15 | 50.67 ± 0.78 | 51.73 ± 0.31 | 51.53 ± 0.20 | 1.89 ± 0.66 | 0.59 ± 0.41 | 1.53 ± 0.34 | 1.34 ± 0.10 | 0.10 ± 0.03 | 0.08 ± 0.01 | 0.14 ± 0.06 | 0.11 ± 0.03 |
| 2 | 1/10 | UFR | 77.87 ± 1.03 | 63.64 ± 1.44 | 52.03 ± 0.47 | 50.58 ± 0.49 | 51.78 ± 0.42 | 51.46 ± 0.29 | 2.29 ± 0.18 | 0.39 ± 0.78 | 2.23 ± 0.27 | 1.64 ± 0.29 | 0.10 ± 0.02 | 0.08 ± 0.03 | 0.11 ± 0.01 | 0.10 ± 0.01 |
| 2 | 1/10⁴ | OSR | 77.29 ± 0.72 | 32.30 ± 0.65 | **51.06** ± 0.52 | 50.32 ± 0.43 | **50.97** ± 0.51 | **50.78** ± 0.35 | 1.05 ± 0.62 | 0.23 ± 0.51 | 1.10 ± 0.55 | 0.79 ± 0.56 | 0.11 ± 0.03 | 0.10 ± 0.04 | 0.11 ± 0.01 | |
| 2 | 1/10⁴ | GSR | 75.61 ± 1.34 | 37.87 ± 3.30 | 51.63 ± 0.16 | **50.09** ± 0.14 | 51.19 ± 0.25 | 50.97 ± 0.12 | 1.87 ± 0.32 | **0.03** ± 0.30 | 1.79 ± 0.62 | 1.23 ± 0.41 | 0.09 ± 0.02 | 0.07 ± 0.03 | **0.07** ± 0.03 | **0.08** ± 0.02 |
| 2 | 1/10⁴ | USR (FDropout) | 76.26 ± 0.95 | 40.61 ± 1.55 | 51.52 ± 0.89 | 50.27 ± 0.14 | 51.19 ± 0.89 | 51.00 ± 0.54 | 2.48 ± 0.05 | 0.16 ± 0.09 | 1.23 ± 0.03 | 1.29 ± 0.02 | **0.08** ± 0.03 | 0.11 ± 0.02 | 0.10 ± 0.05 | 0.10 ± 0.02 |
| 5 | 1 | OFM (HeteroFL) | 78.69 ± 0.59 | 71.00 ± 0.92 | 53.55 ± 0.37 | 51.38 ± 0.45 | 52.77 ± 0.25 | 52.56 ± 0.32 | 3.66 ± 0.62 | 1.87 ± 0.37 | 2.66 ± 0.08 | 2.73 ± 0.36 | 0.13 ± 0.02 | 0.10 ± 0.04 | 0.12 ± 0.04 | 0.12 ± 0.02 |
| 5 | 1 | OFR | 78.84 ± 0.65 | **72.02** ± 0.37 | 53.76 ± 0.39 | 50.93 ± 0.19 | 52.68 ± 0.17 | 52.46 ± 0.07 | 4.24 ± 0.30 | 0.70 ± 0.37 | 3.60 ± 0.67 | 2.85 ± 0.09 | **0.10** ± 0.03 | 0.07 ± 0.09 | **0.07** ± 0.03 | **0.08** ± 0.02 |
| 5 | 1/4 | OSM | **79.00** ± 0.33 | 66.22 ± 1.96 | 52.72 ± 0.60 | 51.10 ± 0.04 | 51.94 ± 0.64 | 51.92 ± 0.41 | 3.38 ± 0.08 | 1.59 ± 0.48 | **1.47** ± 0.18 | 2.15 ± 0.07 | **0.10** ± 0.05 | 0.08 ± 0.01 | 0.13 ± 0.06 | 0.10 ± 0.04 |
| 5 | 1/4 | GFM | 78.89 ± 0.45 | 68.23 ± 0.64 | 53.04 ± 0.06 | 50.89 ± 0.22 | 52.19 ± 0.40 | 52.04 ± 0.18 | 4.11 ± 0.38 | 1.11 ± 0.22 | 2.66 ± 0.19 | 2.62 ± 0.12 | 0.12 ± 0.01 | 0.10 ± 0.03 | 0.09 ± 0.01 | 0.10 ± 0.02 |
| 5 | 1/4 | GFR | 78.95 ± 0.48 | 68.92 ± 0.30 | 52.58 ± 0.43 | **50.76** ± 0.15 | 51.92 ± 0.05 | 51.75 ± 0.18 | **2.55** ± 0.06 | 1.14 ± 0.00 | 1.87 ± 0.53 | **1.85** ± 0.16 | 0.12 ± 0.03 | **0.06** ± 0.02 | 0.10 ± 0.03 | 0.09 ± 0.01 |
| 5 | 1/10 | UFR | 78.83 ± 0.56 | 64.68 ± 0.70 | 52.56 ± 0.44 | 51.04 ± 0.43 | 51.87 ± 0.46 | 51.83 ± 0.18 | 2.65 ± 0.76 | 1.01 ± 0.02 | 2.05 ± 0.83 | 1.90 ± 0.54 | 0.12 ± 0.02 | 0.10 ± 0.04 | 0.08 ± 0.02 | 0.10 ± 0.01 |
| 5 | 1/10⁴ | OSR | 78.86 ± 0.48 | 51.41 ± 3.51 | 52.64 ± 0.22 | 51.23 ± 0.24 | 51.89 ± 0.53 | 51.92 ± 0.17 | 3.41 ± 0.73 | 1.55 ± 0.23 | 2.68 ± 0.62 | 2.55 ± 0.07 | 0.12 ± 0.03 | 0.10 ± 0.04 | 0.08 ± 0.01 | 0.10 ± 0.01 |
| 5 | 1/10⁴ | GSR | 78.50 ± 0.95 | 52.09 ± 1.57 | 52.49 ± 0.20 | 50.89 ± 0.38 | 51.82 ± 0.50 | 51.73 ± 0.03 | 3.12 ± 0.42 | **0.76** ± 0.50 | 1.92 ± 0.83 | 1.93 ± 0.25 | 0.12 ± 0.03 | 0.08 ± 0.03 | 0.09 ± 0.04 | 0.10 ± 0.03 |
| 5 | 1/10⁴ | USR (FDropout) | 78.44 ± 0.38 | 52.31 ± 1.11 | **52.29** ± 0.34 | 51.11 ± 0.49 | **51.67** ± 0.23 | **51.69** ± 0.22 | 2.69 ± 0.58 | 1.00 ± 0.01 | 2.39 ± 0.53 | 2.03 ± 0.37 | 0.13 ± 0.04 | 0.09 ± 0.02 | 0.11 ± 0.04 | 0.11 ± 0.01 |
| 8 | 1 | OFM (HeteroFL) | 79.07 ± 0.42 | **73.96** ± 0.48 | 54.19 ± 0.42 | 51.88 ± 0.15 | 53.10 ± 0.36 | 53.06 ± 0.29 | 3.63 ± 0.28 | 2.03 ± 0.07 | 3.55 ± 0.83 | 3.07 ± 0.39 | 0.13 ± 0.03 | **0.09** ± 0.01 | 0.11 ± 0.04 | 0.11 ± 0.02 |
| 8 | 1 | OFR | 78.69 ± 0.20 | 73.37 ± 0.39 | 53.91 ± 0.26 | 51.76 ± 0.13 | 53.07 ± 0.10 | 52.92 ± 0.11 | 4.18 ± 0.20 | 2.12 ± 0.07 | 2.86 ± 1.39 | 3.05 ± 0.51 | 0.14 ± 0.02 | 0.10 ± 0.04 | 0.12 ± 0.05 | 0.12 ± 0.02 |
| 8 | 1/4 | OSM | 78.85 ± 0.20 | 70.91 ± 2.15 | 53.40 ± 0.33 | 51.96 ± 0.43 | 52.63 ± 0.42 | 52.66 ± 0.15 | **3.06** ± 0.57 | 2.15 ± 0.50 | 2.74 ± 0.50 | **2.65** ± 0.23 | 0.11 ± 0.02 | 0.08 ± 0.01 | **0.07** ± 0.03 | **0.09** ± 0.01 |
| 8 | 1/4 | GFM | 78.82 ± 0.06 | 71.02 ± 0.38 | 53.93 ± 0.22 | **51.50** ± 0.28 | 52.84 ± 0.02 | 52.76 ± 0.16 | 4.01 ± 0.74 | **1.76** ± 0.55 | 3.38 ± 0.87 | 3.05 ± 0.72 | 0.11 ± 0.05 | 0.09 ± 0.03 | 0.12 ± 0.04 | 0.11 ± 0.03 |
| 8 | 1/4 | GFR | 78.64 ± 0.71 | 72.26 ± 1.06 | 53.63 ± 0.47 | 52.07 ± 0.30 | 52.53 ± 0.39 | 52.74 ± 0.27 | 3.21 ± 0.11 | 2.06 ± 0.55 | 2.95 ± 0.53 | 2.74 ± 0.03 | 0.16 ± 0.03 | **0.08** ± 0.02 | 0.16 ± 0.04 | 0.13 ± 0.03 |
| 8 | 1/10 | UFR | **79.10** ± 0.33 | 68.97 ± 1.19 | 53.94 ± 0.72 | 51.79 ± 0.43 | 53.09 ± 0.43 | 52.94 ± 0.44 | 3.92 ± 1.58 | 2.05 ± 1.13 | 3.07 ± 0.47 | 3.01 ± 1.04 | 0.12 ± 0.01 | 0.09 ± 0.03 | 0.12 ± 0.02 | 0.11 ± 0.01 |
| 8 | 1/10⁴ | OSR | 78.82 ± 0.72 | 67.32 ± 0.08 | 53.27 ± 0.35 | 52.19 ± 0.16 | 52.52 ± 0.37 | 52.66 ± 0.27 | 3.33 ± 0.67 | 2.33 ± 0.81 | 2.93 ± 0.98 | 2.86 ± 0.37 | **0.09** ± 0.02 | 0.10 ± 0.05 | 0.11 ± 0.02 | 0.10 ± 0.01 |
| 8 | 1/10⁴ | GSR | 78.38 ± 0.25 | 67.51 ± 0.32 | **53.08** ± 0.25 | 51.68 ± 0.37 | **52.45** ± 0.29 | **52.40** ± 0.27 | 3.74 ± 0.80 | 1.94 ± 0.64 | 2.46 ± 0.12 | 2.71 ± 0.09 | 0.14 ± 0.04 | **0.08** ± 0.00 | 0.12 ± 0.03 | 0.12 ± 0.01 |
| 8 | 1/10⁴ | USR (FDropout) | 78.87 ± 0.30 | 67.74 ± 0.38 | 53.55 ± 0.57 | 51.95 ± 0.25 | 52.72 ± 0.33 | 52.74 ± 0.36 | 3.65 ± 1.62 | 2.18 ± 0.05 | 2.58 ± 0.37 | 2.80 ± 0.68 | 0.13 ± 0.01 | 0.11 ± 0.02 | 0.13 ± 0.03 | 0.13 ± 0.01 |

**TABLE 9.** Detailed results on the CIFAR-100 dataset. Experiments averaged over 3 runs. Best in each category highlighted with bold. Methods are grouped by number of *large* clients and ordered based on the frequency a client receives the same parameters.

| No. of large | Freq. | Method name | ↑ Server Acc % | ↑ Client Acc % | ↓ AUC % tMIA | LiRA | Yeom | Avg | ↓ Attack Adv % tMIA | LiRA | Yeom | Avg | ↓ TPR% at 0.1% FPR tMIA | LiRA | Yeom | Avg |
|---|---|---|---|---|---|---|---|---|---|---|---|---|---|---|---|---|
| 10 | 1 | FedAvg100k | **45.44** ± 0.75 | **44.98** ± 0.52 | 55.20 ± 0.26 | 51.82 ± 0.69 | 54.65 ± 0.12 | 53.89 ± 0.19 | 4.64 ± 0.69 | 1.60 ± 0.80 | 3.10 ± 0.33 | 3.11 ± 0.52 | 0.13 ± 0.05 | 0.10 ± 0.02 | 0.06 ± 0.01 | 0.10 ± 0.02 |
| 0 | 1 | FedAvg30k | 34.01 ± 0.43 | 33.88 ± 0.67 | **52.78** ± 0.12 | **50.91** ± 0.30 | **52.68** ± 0.13 | **52.12** ± 0.03 | **2.89** ± 1.19 | **0.35** ± 1.15 | **1.10** ± 2.17 | **1.45** ± 0.89 | **0.09** ± 0.03 | 0.09 ± 0.01 | **0.05** ± 0.01 | **0.08** ± 0.02 |
| 2 | 1 | OFM (HeteroFL) | 44.14 ± 2.59 | 32.46 ± 0.99 | 54.71 ± 0.19 | 51.15 ± 0.38 | 54.31 ± 0.18 | 53.39 ± 0.19 | 4.07 ± 1.51 | 0.95 ± 0.67 | 3.40 ± 3.11 | 2.89 ± 1.24 | 0.17 ± 0.03 | 0.10 ± 0.05 | 0.07 ± 0.06 | 0.11 ± 0.03 |
| 2 | 1 | OFR | 44.36 ± 1.86 | **33.02** ± 0.72 | 54.62 ± 0.26 | 51.33 ± 0.36 | 54.28 ± 0.17 | 53.41 ± 0.19 | 6.11 ± 2.23 | 1.50 ± 0.49 | 2.43 ± 1.58 | 3.35 ± 0.63 | 0.10 ± 0.01 | **0.07** ± 0.02 | **0.06** ± 0.03 | **0.08** ± 0.02 |
| 2 | 1/4 | OSM | **45.13** ± 2.25 | 22.54 ± 2.16 | 53.06 ± 0.36 | 51.05 ± 0.36 | 52.97 ± 0.42 | 52.36 ± 0.19 | 3.94 ± 2.46 | 0.92 ± 1.81 | 1.70 ± 1.79 | 2.19 ± 1.60 | 0.10 ± 0.04 | 0.09 ± 0.02 | 0.09 ± 0.05 | 0.09 ± 0.04 |
| 2 | 1/4 | GFM | 43.05 ± 1.76 | 31.70 ± 1.37 | 53.11 ± 0.62 | 50.96 ± 0.49 | 52.95 ± 0.38 | 52.34 ± 0.31 | 4.24 ± 1.58 | 0.50 ± 0.85 | **0.88** ± 2.32 | 1.88 ± 0.14 | 0.08 ± 0.02 | 0.11 ± 0.04 | 0.10 ± 0.03 | |
| 2 | 1/4 | GFR | 43.17 ± 2.74 | 30.44 ± 0.50 | 53.08 ± 0.39 | 51.10 ± 0.22 | 53.08 ± 0.23 | 52.42 ± 0.24 | 4.47 ± 0.61 | 1.27 ± 0.47 | 2.13 ± 1.55 | 2.62 ± 0.47 | 0.12 ± 0.03 | 0.09 ± 0.04 | 0.09 ± 0.01 | 0.10 ± 0.02 |
| 2 | 1/10 | UFR | 43.16 ± 1.83 | 26.58 ± 1.38 | 52.66 ± 0.56 | 50.79 ± 0.13 | 52.73 ± 0.48 | 52.06 ± 0.30 | 2.68 ± 1.70 | **-0.23** ± 1.61 | 1.29 ± 1.29 | **1.25** ± 0.48 | 0.12 ± 0.04 | 0.11 ± 0.02 | 0.07 ± 0.02 | 0.10 ± 0.01 |
| 2 | 1/10⁴ | OSR | 43.05 ± 1.53 | 10.85 ± 0.28 | 52.28 ± 0.22 | 50.87 ± 0.20 | 52.11 ± 0.12 | 51.75 ± 0.09 | 2.86 ± 1.54 | 0.75 ± 1.90 | 0.98 ± 3.26 | 1.53 ± 0.28 | 0.11 ± 0.06 | 0.08 ± 0.04 | 0.14 ± 0.10 | 0.11 ± 0.04 |
| 2 | 1/10⁴ | GSR | 42.32 ± 1.99 | 12.51 ± 0.94 | **52.12** ± 0.70 | 50.68 ± 0.15 | **51.89** ± 0.61 | **51.56** ± 0.40 | 2.24 ± 2.40 | 0.99 ± 0.71 | 1.08 ± 2.60 | 1.43 ± 0.46 | 0.10 ± 0.03 | **0.07** ± 0.03 | 0.10 ± 0.03 | 0.09 ± 0.03 |
| 2 | 1/10⁴ | USR (FDropout) | 41.31 ± 1.67 | 12.85 ± 0.62 | 52.36 ± 0.62 | **50.54** ± 0.10 | 52.03 ± 0.35 | 51.65 ± 0.23 | 3.41 ± 2.31 | 1.51 ± 0.31 | 1.15 ± 0.78 | 2.02 ± 0.91 | **0.09** ± 0.02 | **0.07** ± 0.02 | 0.10 ± 0.02 | 0.09 ± 0.01 |
| 5 | 1 | OFM (HeteroFL) | 44.98 ± 1.22 | 35.89 ± 0.24 | 55.23 ± 0.31 | 51.58 ± 0.09 | 54.75 ± 0.12 | 53.85 ± 0.14 | 4.87 ± 2.11 | 1.89 ± 0.87 | 3.75 ± 1.41 | 3.50 ± 1.06 | 0.14 ± 0.03 | 0.09 ± 0.04 | 0.05 ± 0.01 | 0.10 ± 0.01 |
| 5 | 1 | OFR | 45.39 ± 1.33 | **36.08** ± 0.36 | 54.82 ± 0.16 | 51.34 ± 0.26 | 54.50 ± 0.16 | 53.55 ± 0.06 | 5.11 ± 1.48 | 1.06 ± 0.61 | **1.31** ± 2.56 | 2.49 ± 0.83 | 0.11 ± 0.05 | **0.08** ± 0.02 | 0.09 ± 0.02 | **0.09** ± 0.02 |
| 5 | 1/4 | OSM | **45.70** ± 1.24 | 29.81 ± 0.66 | 53.97 ± 0.46 | 51.14 ± 0.66 | 53.69 ± 0.25 | 52.93 ± 0.36 | 4.85 ± 0.43 | **-0.36** ± 0.84 | 2.49 ± 1.66 | 2.33 ± 0.75 | 0.13 ± 0.02 | 0.11 ± 0.02 | **0.06** ± 0.02 | 0.10 ± 0.01 |
| 5 | 1/4 | GFM | 44.57 ± 0.22 | 32.49 ± 0.66 | 53.92 ± 0.02 | 51.38 ± 0.31 | 53.59 ± 0.05 | 52.96 ± 0.11 | 4.82 ± 1.32 | 1.49 ± 0.61 | 2.50 ± 1.53 | 2.94 ± 0.83 | 0.10 ± 0.03 | **0.07** ± 0.03 | 0.11 ± 0.03 | 0.10 ± 0.01 |
| 5 | 1/4 | GFR | 45.22 ± 1.00 | 32.81 ± 0.37 | 54.04 ± 0.08 | 51.61 ± 0.37 | 53.77 ± 0.15 | 53.14 ± 0.18 | 4.04 ± 0.89 | 0.87 ± 0.62 | 1.62 ± 1.54 | 2.18 ± 0.85 | **0.10** ± 0.01 | 0.12 ± 0.04 | 0.07 ± 0.03 | 0.10 ± 0.00 |
| 5 | 1/10 | UFR | 44.98 ± 0.09 | 29.36 ± 1.24 | 53.78 ± 0.27 | 51.19 ± 0.09 | 53.57 ± 0.29 | 52.85 ± 0.16 | 5.52 ± 1.74 | 0.94 ± 0.13 | 2.78 ± 1.10 | 3.08 ± 0.88 | 0.12 ± 0.05 | 0.10 ± 0.02 | 0.08 ± 0.00 | 0.10 ± 0.01 |
| 5 | 1/10⁴ | OSR | 44.92 ± 1.22 | 24.27 ± 0.54 | 53.76 ± 0.73 | 50.88 ± 0.32 | 53.19 ± 0.47 | 52.61 ± 0.37 | 5.02 ± 0.81 | 1.38 ± 0.81 | 2.80 ± 0.79 | 3.07 ± 0.70 | 0.12 ± 0.03 | 0.09 ± 0.06 | 0.11 ± 0.03 | 0.11 ± 0.03 |
| 5 | 1/10⁴ | GSR | 44.84 ± 0.89 | 24.20 ± 0.40 | **53.56** ± 0.34 | **50.81** ± 0.24 | 53.24 ± 0.12 | **52.54** ± 0.27 | 3.98 ± 0.46 | 1.41 ± 0.44 | 2.09 ± 1.72 | 2.49 ± 0.49 | 0.10 ± 0.03 | 0.10 ± 0.02 | 0.07 ± 0.02 | **0.09** ± 0.02 |
| 5 | 1/10⁴ | USR (FDropout) | 45.52 ± 0.60 | 24.41 ± 0.10 | 53.73 ± 0.08 | 51.16 ± 0.18 | **53.06** ± 0.09 | 52.65 ± 0.12 | 4.04 ± 1.25 | 1.13 ± 1.92 | 1.96 ± 1.90 | 2.38 ± 1.54 | **0.10** ± 0.01 | 0.12 ± 0.06 | 0.07 ± 0.04 | 0.10 ± 0.01 |
| 8 | 1 | OFM (HeteroFL) | 44.85 ± 1.32 | 39.02 ± 1.43 | 54.95 ± 0.26 | 51.59 ± 0.34 | 54.61 ± 0.23 | 53.72 ± 0.21 | 6.01 ± 1.22 | 2.40 ± 0.67 | 2.17 ± 0.27 | 3.53 ± 0.64 | 0.09 ± 0.04 | 0.10 ± 0.01 | 0.07 ± 0.04 | 0.09 ± 0.03 |
| 8 | 1 | OFR | 45.71 ± 0.99 | **40.06** ± 0.67 | 55.18 ± 0.35 | 51.63 ± 0.10 | 54.88 ± 0.07 | 53.90 ± 0.07 | 5.09 ± 1.22 | 1.37 ± 1.31 | 2.81 ± 1.03 | 3.11 ± 1.08 | 0.11 ± 0.06 | 0.15 ± 0.03 | 0.10 ± 0.01 | 0.12 ± 0.00 |
| 8 | 1/4 | OSM | 45.97 ± 0.62 | 37.98 ± 0.28 | 54.33 ± 0.34 | 51.59 ± 0.43 | 54.05 ± 0.02 | 53.32 ± 0.30 | 4.09 ± 0.49 | 1.17 ± 1.53 | 1.22 ± 2.29 | 2.16 ± 0.19 | 0.13 ± 0.02 | 0.09 ± 0.00 | **0.07** ± 0.02 | 0.10 ± 0.01 |
| 8 | 1/4 | GFM | 45.55 ± 0.76 | 37.98 ± 1.28 | 54.49 ± 0.27 | 51.56 ± 0.22 | 54.01 ± 0.14 | 53.35 ± 0.08 | 6.14 ± 1.10 | 1.91 ± 1.19 | **1.02** ± 1.62 | 3.02 ± 0.29 | 0.12 ± 0.05 | 0.09 ± 0.03 | **0.07** ± 0.02 | 0.09 ± 0.01 |
| 8 | 1/4 | GFR | 46.30 ± 0.54 | 38.44 ± 0.66 | 54.66 ± 0.26 | 51.55 ± 0.48 | 54.17 ± 0.14 | 53.46 ± 0.09 | 5.68 ± 1.43 | 1.72 ± 0.48 | 3.04 ± 0.42 | 3.48 ± 0.38 | 0.13 ± 0.05 | 0.09 ± 0.01 | **0.07** ± 0.01 | 0.10 ± 0.02 |
| 8 | 1/10 | UFR | **46.54** ± 0.69 | 38.03 ± 0.39 | 54.50 ± 0.13 | 51.74 ± 0.47 | 54.06 ± 0.20 | 53.43 ± 0.06 | 5.46 ± 0.31 | 1.22 ± 0.42 | 2.98 ± 1.45 | 3.22 ± 0.15 | 0.11 ± 0.01 | **0.08** ± 0.02 | 0.09 ± 0.02 | 0.09 ± 0.01 |
| 8 | 1/10⁴ | OSR | 45.30 ± 0.47 | 37.03 ± 0.25 | **54.00** ± 0.20 | 51.68 ± 0.25 | 53.68 ± 0.05 | 53.12 ± 0.16 | 4.56 ± 1.31 | 1.97 ± 0.47 | 2.73 ± 0.73 | 3.09 ± 0.82 | 0.19 ± 0.00 | 0.10 ± 0.04 | 0.12 ± 0.07 | 0.13 ± 0.01 |
| 8 | 1/10⁴ | GSR | 45.65 ± 0.42 | 36.81 ± 0.59 | 54.03 ± 0.21 | 51.74 ± 0.45 | **53.68** ± 0.34 | **53.15** ± 0.33 | 3.72 ± 0.66 | **0.56** ± 0.95 | 1.27 ± 1.01 | **1.85** ± 0.61 | 0.14 ± 0.02 | 0.09 ± 0.04 | **0.07** ± 0.02 | 0.10 ± 0.02 |
| 8 | 1/10⁴ | USR (FDropout) | 45.55 ± 0.44 | 37.14 ± 0.28 | 53.77 ± 0.35 | **51.48** ± 0.36 | 53.47 ± 0.54 | 52.91 ± 0.41 | 4.45 ± 0.23 | 1.26 ± 0.50 | 2.19 ± 2.23 | 2.63 ± 0.66 | 0.12 ± 0.03 | 0.09 ± 0.05 | 0.13 ± 0.05 | 0.12 ± 0.01 |

integrated two commonly used methods to tackle non-IIDness in federating learning, namely `FedProx` [44], and `FedAvgM` [45]. `FedProx` applies a proximal term in the client loss based on the client model's distance from the server model, which regularizes the client model updates, reducing the differences between client updates and therefore improving client privacy. `FedAvgM` uses momentum on the server side. It has been shown to be a simple yet effective method to improve performance in non-IID settings. Table 6 summarizes the performance of all the heterogeneous FL approaches with non-IID data after applying `FedProx` and `FedAvgM`. As seen in the Table, there is an increase in the server and client accuracy and an improvement in client privacy for most of the heterogeneous FL methods. In addition, other approaches that have been proposed in the literature to mitigate the challenges of non-IID data could also be used, including contrasive loss [46], [47], distillation [48], and representation learing [49]. We leave to future work a more in-depth exploration of such methods as it is out of the main scope of this paper.

In future work we also plan to further study the behavior of model-agnostic FL methods on non-IID data. We speculate that these differences in behavior might be due to spurious correlations, which are inexistent in the CIFAR-x dasets yet present in the FEMNIST dataset, as the writer's style might be correlated to certain classes and clients [40].

### C. ATTACKS ON CIFAR-10 AND CIFAR-100 DATASETS

Results on the two CIFAR datasets are summarized in the main paper in included in detail in Table 8 and Table 9.

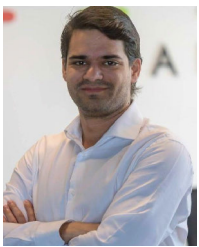

**GERGELY D. NÉMETH** received the degree in computer science from Budapest University of Technology and Economics, in 2018. and the degree in computer science from The University of Manchester, in 2019. He is an ELLIS Ph.D. student at ELLIS Alicante and the University of Alicante.

He involved on natural language processing (NLP) at The University of Manchester. Later, he involved on applied machine learning (ML) projects focused on computer vision (CV) at Asura Technologies, a Hungarian StartUp leading in smart traffic applications. He focuses on the human interest in the cooperative learning systems. His supervisors are Nuria Oliver (ELLIS Alicante), Miguel Angel Lozano (University of Alicante), and Novi Quadrianto (University of Sussex).

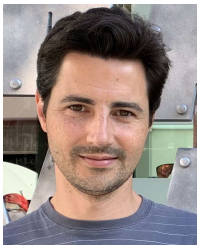

**MIGUEL ÁNGEL LOZANO** received the degree and Ph.D. degrees in computer engineering from the University of Alicante, in 2001 and 2008, respectively. In 2008, he defended his Ph.D. Thesis, entitled "Kernelized Graph Matching and Clustering."

In 2002, he obtained a FPI research grant, and since 2004, he has been a Lecturer with the Department of Computer Science and Artificial Intelligence, University of Alicante. He has conducted research stays at the Computer Vision and Pattern Recognition Laboratory, University of York, and the Bioinformatics Laboratory, University of Helsinki. From 2002 to 2010, he developed his research within the Robot Vision Group (RVG), and in 2010, he joined the Mobile Vision Research Laboratory (MVRLab). He is currently the Head of MVRLab, a group that focuses on pattern recognition and computer vision on mobile devices, the Director of the master's degree in software development for mobile devices, and the Coordinator for Quality Assurance and Educational Innovation at the Polytechnic School. His research interests include pattern recognition, graph matching and clustering, and computer vision. Within MVRLab and in collaboration with Neosistec, some applications aimed to assist people with limited vision have been developed: aerial obstacle detection (AOD), SuperVision, and NaviLens. AOD and NaviLens won the "Application Mobile for Good" Award in the 7th and 11th Vodafone Foundation Awards, in 2014 and 2017, respectively. In 2020, he joined the Data Science against COVID19 taskforce of the Valencian Government. This team participated and won first prize in the XPRIZE Pandemic Response Challenge. Moreover, he is tutoring two ELLIS Ph.D. students at ELLIS Alicante.

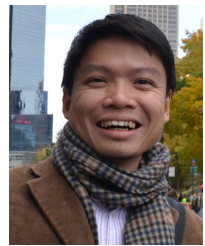

**NOVI QUADRIANTO** received the Ph.D. degree in computer science from The Australian National University, Canberra, Australia, in 2012. Since March 2021, he has been leading the BCAM Severo Ochoa Strategic Laboratory on Trustworthy Machine Learning, Spain. Since July 2022, he has been holding an adjunct position at Monash University, Indonesia. He is currently a Professor of machine learning with the University of Sussex, U.K. Prior to Sussex, he was a Newton International Fellow of the Royal Society and the British Academy, Machine Learning Group, Department of Engineering, University of Cambridge. In 2019, he was awarded an European Research Council ERC Starting Grant for a project on developing Bayesian models and algorithms for fairness and transparency (BayesianGDPR).

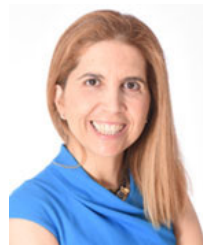

**NURIA OLIVER** (Fellow, IEEE) received the Ph.D. degree from the Media Laboratory, MIT, and the Honorary Doctorate from the University Miguel Hernández.

She is currently the Scientific Director and one of the founders of ELLIS Alicante. She is the Co-Founder and the Vice-President of ELLIS Alicante. Previously, she was the Chief Scientific Advisor of the Vodafone Institute, the Director of the Data Science Research, Vodafone, the Scientific Director of Telefónica, and a Researcher with Microsoft Research. She is also the Chief Data Scientist of DataPop Alliance. During the COVID-19 pandemic, she held an honorary position as the Commissioner to the President of the Valencian Government on AI and Data Science against COVID-19. She advises several universities, governments, and companies.

Dr. Oliver is an ACM, ELLIS, and EurAI Fellow, and an Elected Permanent Member of the Royal Academy of Engineering of Spain. She is also a member of the CHI Academy and the Academia Europaea and a Corresponding Member of the Academy of Engineering of Mexico. She is well known for her work in computational models of human behavior, human–computer-interaction, mobile computing, and big data for social good. Named inventor of 40 patents. She has received many awards, including the MIT TR100 Young Innovator Award (2004), Spanish National Computer Science Award (2016), the Engineer of the Year (2018), the Valencian Medal to Business and Social Impact (2018), the Data Scientist of the Year (2019), the Jaume I Award in New Technologies (2021), the Abie Technology Leadership Award by AnitaB.org (2021), and the Hipatia European Award to Excellence in Science (2024). According to Research.com, she is Spanish female computer scientist with largest scientific impact (2022–2024).

○ ○ ○